# Computational Aspects of Reordering Plans

**Christer Bäckström**                                        CBA@IDA.LIU.SE
*Department of Computer and Information Science*
*Linköpings universitet, S-581 83 Linköping, Sweden*

## Abstract

This article studies the problem of modifying the action ordering of a plan in order to optimise the plan according to various criteria. One of these criteria is to make a plan less constrained and the other is to minimize its parallel execution time. Three candidate definitions are proposed for the first of these criteria, constituting a sequence of increasing optimality guarantees. Two of these are based on deordering plans, which means that ordering relations may only be removed, not added, while the third one uses reordering, where arbitrary modifications to the ordering are allowed. It is shown that only the weakest one of the three criteria is tractable to achieve, the other two being NP-hard and even difficult to approximate. Similarly, optimising the parallel execution time of a plan is studied both for deordering and reordering of plans. In the general case, both of these computations are NP-hard. However, it is shown that optimal deorderings can be computed in polynomial time for a class of planning languages based on the notions of producers, consumers and threats, which includes most of the commonly used planning languages. Computing optimal reorderings can potentially lead to even faster parallel executions, but this problem remains NP-hard and difficult to approximate even under quite severe restrictions.

## 1. Introduction

In many applications where plans, made by man or by computer, are executed, it is important to find plans that are optimal with respect to some cost measure, typically execution time. Examples of such applications are manufacturing and error-recovery for industrial processes, production planning, logistics and robotics. Many different kinds of computations can be made to improve the cost of a plan—only a few of which have been extensively studied in the literature. The most well-known and frequently used of these is *scheduling*. A plan tells which actions (or tasks) to do and in which order to do them, while a schedule assigns exact release times to these actions. The schedule must obey the action order prescribed by the plan and must often also satisfy further metric constraints such as deadlines and earliest release times for certain actions. A schedule is *feasible* if it satisfies all such metric constraints. It is usually interesting to find a schedule that is optimal in some respect, *eg* the feasible schedule having the shortest total execution time, or the schedule missing the deadlines for as few actions as possible.

In principle, planning and scheduling follow in sequence such that scheduling can be viewed as a post-processing step to planning—where planning is concerned with causal relations and qualitative temporal relations between actions, while scheduling is concerned with metric constraints on actions. In some planning systems, *eg* O-PLAN (Currie & Tate, 1991) and SIPE (Wilkins, 1988), both planning and scheduling are integrated into one single system. Similarly, temporal planners, *eg* DEVISER (Vere, 1983) and IxTeT (Ghallab & Laruelle, 1994), can often reason also about metric constraints. This does not make it irrelevant to study planning and scheduling as separate problems, though, as can be seen from the vast literature on both topics. The two problems are of quite different character and studying them separately gives important insight also into such integrated systems as was just discussed. For instance, Drabble[1] says that it is often very difficult to see when O-PLAN plans and when it schedules; it is easy to see that O-PLAN works, but it is difficult to see why.

A further complication in understanding the difference between planning and scheduling, both for integrated systems and for systems with separated planning and scheduling, is that certain types of computations fall into a grey zone between planning and scheduling. Planners are good at reasoning about effects of actions and causal relationships between actions, but are usually very poor at reasoning about time and temporal relationships between actions. Schedulers, on the other hand, are primarily designed to reason about time and resource conflicts, but have no capabilities for reasoning about causal dependencies between actions. The problems in the grey zone require reasoning of both kinds, so neither planners nor schedulers can handle these problems properly. If these problems are not solved, then the scheduler does not get sufficient information from the planner to do the best of the situation—the planner and the scheduler may fail in their cooperation to find a plan with a feasible schedule, even when such a plan exists.

This article focusses on one of these grey-zone problems, namely the problem of optimising the action order of a plan to allow for better schedules. Whenever two actions conflict with each other and cannot be allowed to execute in parallel, a planner must order these actions. However, it usually does not have enough information and reasoning capabilities to decide which of the two possible orders is the best one, so it makes an arbitrary choice. One of the choices typically allows for a better schedule than the other one, so if the planner makes the wrong choice it may prevent the scheduler from finding a good, or even feasible, schedule. This situation arises also when plans are made by a human expert, since it is difficult to see which choice of ordering is the best one in a large and complex plan. Planning systems of today usually cannot do anything better than asking the planner for a new plan if the scheduler fails to find a feasible schedule. This is an expensive and unsatisfactory solution, especially if there is no feedback from the scheduler to help the planner making a more intelligent choice next time. Another solution which appears in the literature is to use a filter between the planner and scheduler which attempts to modify the plan order to put the scheduler in a better position. Such filters could remove certain over-commitments in the ordering, which will be referred to as *deordering* the plan, or even change the order between certain actions, which will be referred to as *reordering* the plan.

This article is intended to provide a first formal foundation for studying this type of problems. It defines a number of different optimality criteria for plan order modifications, both with respect to the degree of over-committment in the ordering and with respect to the parallel execution time, and it also provides computational results for computing such modifications. The article also analyses some filtering algorithms suggested in the literature for doing such order modifications.

The remainder of this article is structured as follows. Section 2 introduces the concepts and computations studied in this article by means of an example. Then Section 3 starts the theoretical content of the article, defining the two planning formalisms used in the following sections. The problems of making a plan least-constrained are studied in Section 4 where some candidate definitions for this concept are introduced and their computational properties investigated. Section 5 defines the concepts of parallel plans and parallel executions of plans. This is followed by Section 6 where optimal deorderings and reorderings of parallel plans are introduced and the complexity of achieving such optimality is analysed. Section 7 then studies how the complexity of these problems is affected by restricting the language. This includes the positive result that an algorithm from the literature finds optimal deorderings for a class of plans for most common planning languages. Some other filtering algorithms from the literature as well as some planners incorporating some ordering optimisation are discussed in Section 8. Finally, Section 9 discusses some aspects of this article and some related work, while Section 10 concludes by a brief recapitulation of the results.

---

1. Brian Drabble, personal communication, Aug. 1997.

## 2. Example

In order to illustrate the concepts and operations studied in this article a simple example of assembling a toy car will be used. The example is a variation of the example used by Bäckström and Klein (1991), which is a much simplified version of an existing assembly line for toy cars used for undergraduate laborations in digital control at Linköping University (for a description of this assembly line, see *eg.* Klein, Jonsson, & Bäckström, 1995, 1998; Strömberg, 1991). The problem is to assemble a LEGO[2] car from pre-assembled parts as shown in Figure 1. There is a chassis, a top and a set of wheels, the two latter to be mounted onto the chassis.

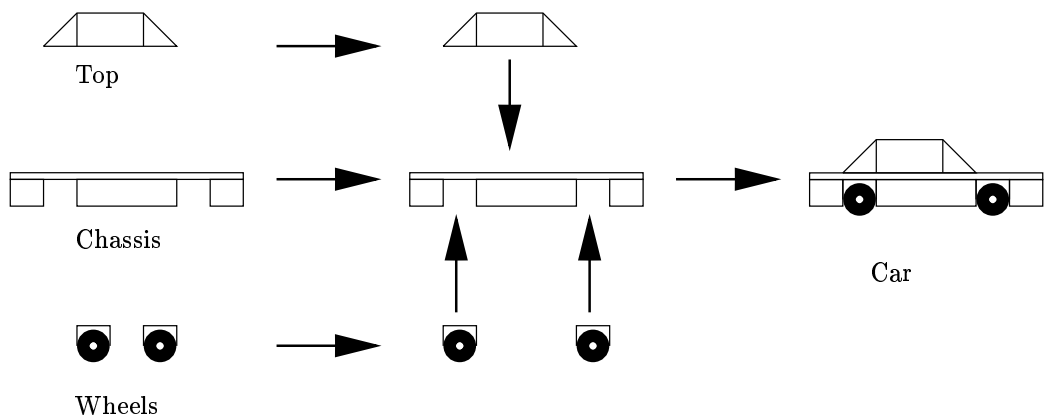

Figure 1: Schematic assembly process for a toy car

The workpiece flow of the factory is shown in Figure 2. There are three storages, one for each type of preassembled part, two workstations, number 1 for mounting the top and number 2 for mounting the wheels, and there is a car storage for assembled cars. Tops can be moved from the top storage to workstation 1 and sets of wheels can be moved from the wheels storage to workstation 2. Chassis can be moved from the chassis storage to either workstation and also, possibly with other parts mounted, between the two workstations and from either workstation to the car storage. Furthermore, before mounting the wheels on a chassis, the tyres must be inflated, so workstation 2 incorporates a compressed-air container which must be pressurized before inflating the tyres (this is not shown in the figure).

---

2. LEGO is a trade mark of the LEGO company

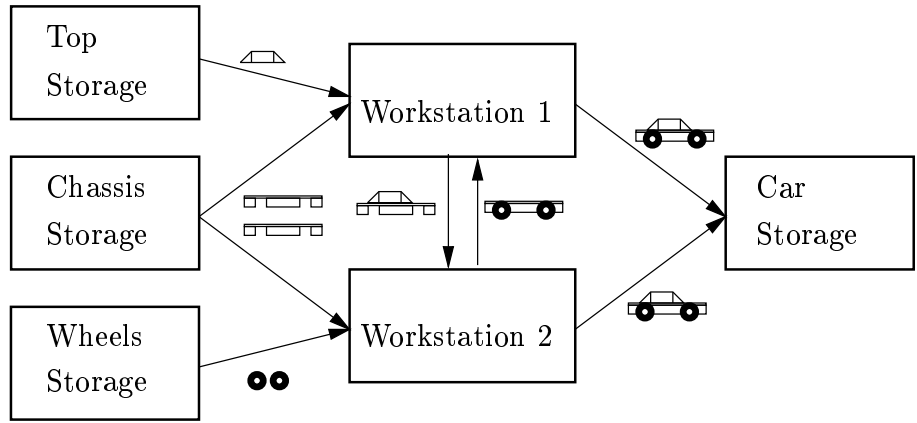

Figure 2: Schematic lay-out of the toy-car factory

This article is concerned with modifying the order between the actions in a given plan, and does not consider modifying also the set of actions. Hence, the example will assume that a plan for assembling a toy car is given—whether this plan was produced by hand or by a planning algorithm is not important. It will also be assumed that this assembly plan contains exactly those actions listed in Table 1, in some order. Since most results in this article are independent of the particular planning language used, no assumptions about the planning language will be made in this example either. To make things simple, the obvious common-sense constraints on which plans are valid will be used. For instance, a part must be moved to a workstation before it is mounted there, the wheels must be inflated before being mounted and the air container must be pressurized before inflating the tyres. Furthermore, since a chassis can only be at one single place at a time, the top cannot be mounted in parallel with mounting the wheels, and neither of the mounting operations can be done in parallel with moving either the chassis or the part to be mounted.

The purpose of modifying the action order in a given plan is usually to optimize the plan in some aspect, for instance, to make the plan *least constrained*. Consider the totally ordered plan in Figure 3a, for producing a chassis with wheels, which is a subplan of the plan for assembling a car. Note that since the plan is totally ordered, all pairs of actions are ordered, but the implicit transitive arcs are not shown in the figure. This plan is clearly over-constrained. For instance, it is not necessary to move the set of wheels to workstation 2 before pressurizing the air container, and removing this ordering constraint results in the plan in Figure 3b. Note that orderings have only been removed—the arc from MvW2 to IT existed already in the original plan, but was implicit by transitivity. A plan where some orderings have been removed will be referred to as a *deordering* of the original plan.

| Action | Description | Duration |
|--------|-------------|----------|
| MvT1 | Move top to workstation 1 | 1 |
| MvW2 | Move wheels to workstation 2 | 1 |
| MvC1 | Move chassis to workstation 1 | 2 |
| MvC2 | Move chassis to workstation 2 | 2 |
| MvS | Move chassis to car storage | 3 |
| MtT | Mount top on chassis | 7 |
| MtW | Mount wheels on chassis | 4 |
| PAC | Pressurize air container | 5 |
| IT | Inflate tyres | 4 |

Table 1: Actions of the assembly plan

This new plan is less constrained than the original plan, since it is now possible to move the wheels and pressurize the air container in either order or, perhaps, even in parallel. However, further orderings can be removed; it is not necessary to inflate the wheels before moving the chassis to the workstation. Removing also this ordering results in the plan in Figure 3c, which is a least constrained deordering of the original plan in the sense that it is not possible to remove any further ordering constraints and still have a valid plan. That is, if removing any further ordering constraint, it will be possible to sequence the actions in such a way that the plan will no longer have its intended result. In addition to deorderings, one may also consider arbitrary modifications of the ordering relation, that is, both removing and adding relations. Such modifications will be referred to as *reorderings*. Three differents least-constraintment criteria for plans based on deorderings and reorderings will be studied in Section 4, and the plan in Figure 3c happens to be optimal according to all three of these criteria.

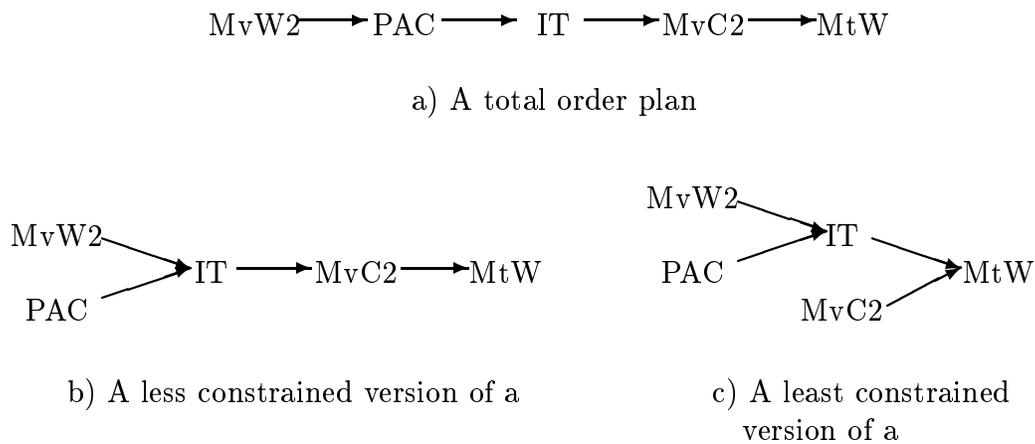

a) A total order plan

b) A less constrained version of a

c) A least constrained version of a

Figure 3: Three plans for mounting the wheels

Making a plan least constrained is clearly useful if certain actions can be executed in parallel. However, even in the case where no parallel execution is possible, it may still be worth making a plan least constrained. Although the partial order of this least constrained plan must again be strengthened into a total order for execution purposes, this need not be the same total order as in the original plan. Suppose the actions have temporal constaints like deadlines and earliest release times and that a scheduler will post-process the plan to try finding a feasible schedule. It may then be the case that the original plan has no feasible schedule, but a less constrained version of it can be sequenced into a feasible schedule. The idea of a least constrained plan is that the scheduler will have as many alternative execution sequences as possible to choose from.

The most important reason for modifying the action ordering of a plan, however, is to execute the plan faster by executing actions in parallel whenever possible. For this purpose it is better to use the length of the optimal schedule for a plan as a measure, rather than some measure on the ordering itself. Suppose the following car-assembly plan is given

$$\langle MvW2, PAC, IT, MvC2, MtW, MvT1, MvC1, MtT, MvS \rangle.$$

If the actions are executed sequentially in the given order, the minimum execution time is the sum of the durations of the actions, that is 29 time units. However, just as in the previous example this plan is over-constrained, since several of the actions could be executed in either order, or in parallel.

It is possible to remove orderings as far as shown in Figure 4a, but no further, and still have a valid plan (the implicit transitive orderings are not shown in the figure). This deordered version of the original assembly plan can be scheduled to execute in 25 time units by exploiting parallelism whenever possible. An example of such a schedule is shown in Figure 3b. However, no faster execution is possible, since the plan contains a subsequence of actions which cannot be parallelized and which has a total execution time of 25 time units.

It is obvious from the schedule in Figure 4b that not many actions can be executed in parallel, and that the gain of deordering the plan is quite small. A much better performance is possible if arbitrary modifications to the action ordering are allowed, that is, if also reorderings are considered. For instance, in the assembly plan there is no particular reason why the wheels should be mounted before the top is mounted, and it will be seen shortly that much time can be saved by reversing the order of these two operations. A deordering cannot do this, however, since removing the ordering between the wheel-mounting action (MtW) and the top-mounting action (MtT) would make these unordered. This would be interpreted as if the two actions could be executed in parallel, which is not possible. This is also the reason why these actions must be ordered in the original plan. However, when allowing arbitrary modifications, the order between these two actions can be reversed, and Figure 5a shows such a reordering of the original plan. This plan can be scheduled to execute in only 16 time units, which is a considerable improvement over both the original plan and the optimal deordered version of it. An example of an optimal schedule is shown in Figure 5b. In fact, this plan is an optimal reordering in the sense that no other ordering of the actions results in a valid plan that can be scheduled to execute faster. The problems of finding optimal deorderings and reorderings of plan with respect to parallel execution is the main topic of this article, and are studied in Sections 5 to 7.

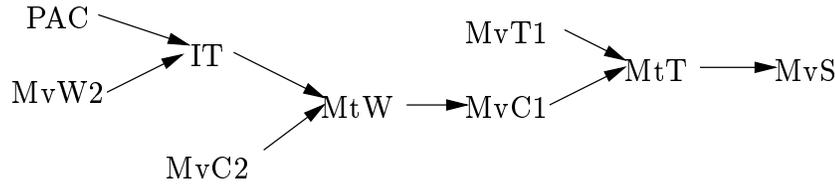

a) A deordering of the assembly plan admitting a shortest parallel execution time

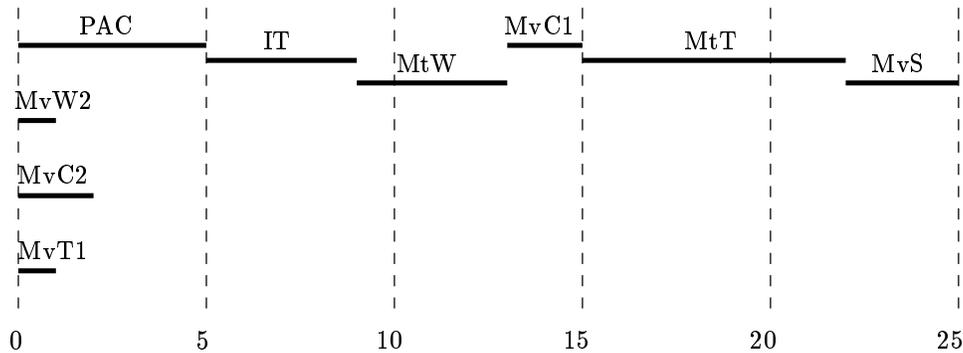

b) An optimal schedule for the plan above

Figure 4: An optimal deordering of the assembly plan

It is obvious that reordering is a more powerful operation than deordering, since the reordered plan in Figure 5a allows for a shorter schedule than the optimal deordering in Figure 4a. On the other hand, if the original plan had been

$$\langle MvT1, MvC1, MtT, MvS, MvW2, PAC, IT, MvC2, MtW \rangle,$$

then deordering would have been sufficient for arriving at the optimal plan in Figure 5a.

## 3. Planning Formalisms

This section defines actions, plans and related concepts, which basically appear in two different guises in this article. Definitions and tractability results will mostly be cast in a general, axiomatic framework in order to be as general and independent of formalism as possible. Hardness results, on the other hand, will mostly be cast in a specific formalism, GROUND TWEAK, and often subject to further restrictions, this in order to strengthen the results. Both these formalisms are defined below. In addition to these, a third formalism will be used, but its definition will be deferred until it is used, in Section 7.

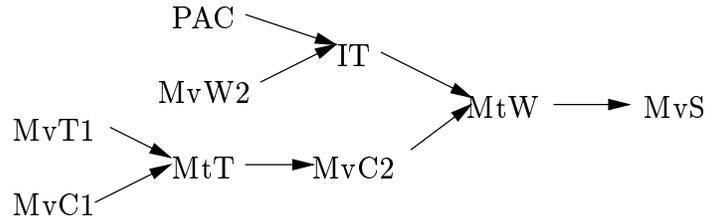

a) A reordering of the assembly plan admitting a shortest
   parallel execution time

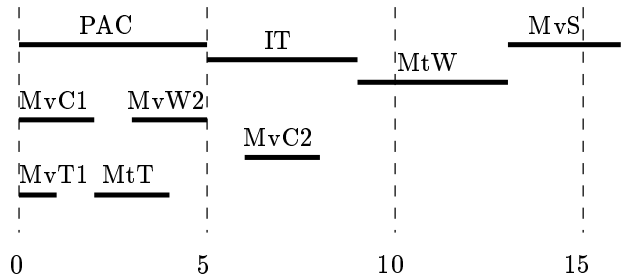

b) An optimal schedule for the plan above

Figure 5: An optimal reordering of the assembly plan

## 3.1 The Axiomatic Planning Framework

The axiomatic framework makes only a minimum of assumptions about the underlying formalism. It may be instantiated to any planning formalism that defines some concept of a *planning problem* a domain of entities called *actions* and a *validity* test. The planning problem is assumed to consist of *planning problem instances* (PPIs),[3] with no further assumptions about the inner structure of these. The validity test is a truth-valued function taking a PPI and a sequence of actions as arguments. If the validity test is true for a PPI $\Pi$ and an action sequence $\langle a_1, \ldots, a_n \rangle$, then the action sequence $\langle a_1, \ldots, a_n \rangle$ is said to *solve* $\Pi$. While the inner structure of the PPIs and the exact definition of the validity test are crucial for any specific planning formalism, many results in this article can be proven without making any such further assumptions. Results on the computational complexity of certain problems will make an assumption about the complexity of the validity test, though. Based on these concepts, the notion of plans can be defined in the usual way.

**Definition 3.1** *A* total-order plan (t.o. plan) *is a sequence* $P = \langle a_1, \ldots, a_n \rangle$ *of actions, which can alternatively be denoted by the tuple* $\langle \{a_1, \ldots, a_n\}, \prec \rangle$ *where for* $1 \leq k, l \leq n$, $a_k \prec a_l$ *iff* $k < l$. *Given a* PPI $\Pi$, $P$ *is said to be* $\Pi$-valid *iff the validity test is true for* $\Pi$ *and* $P$.

---

3. This is the complexity-theoretic terminology for problems. Planning problem instances in the sense of this article are sometimes referred to as planning problems in the planning literature.

*A* partial-order plan (p.o. plan) *is a tuple* $P = \langle A, \prec \rangle$ *where A is a set of actions and* $\prec$ *is a strict ( ie. irreflexive) partial order on A. The validity test is extended to p.o. plans s.t. given a* PPI $\Pi$, *P is* $\Pi$-*valid iff* $\langle A, \prec' \rangle$ *is valid for every topological sorting* $\prec'$ *of* $\prec$.

The actions of a t.o. plan must be executed in the specified order, while unordered actions in a p.o. plan may be executed in either order. That is, a p.o. plan can be viewed as a compact representation for a set of t.o. plans. There is no implicit assumption that unordered actions can be executed in parallel; parallel plans will be defined in Section 5. p.o. plans will be viewed as *directed acyclic graphs* in figures with the transitive arcs often tacitly omitted to enhance readability. Furthermore, all proofs and algorithms in this article are based on this definition, *ie* assuming the order of a plan is transitively closed, while many practical planners do not bother about transitive closures. This difference does not affect any of the results presented here.

## 3.2 The Ground TWEAK Formalism

The Ground TWEAK (GT) formalism is the TWEAK language (Chapman, 1987) restricted to ground actions. This formalism is a variation on propositional STRIPS and it is known to be equivalent under polynomial transformation to most other common variants on propositional STRIPS (Bäckström, 1995). In brief, an action has a precondition and a postcondition, both being sets of ground literals.

In order to define the GT formalism, the following two definitions are required. Given some set $S$, the notion $Seqs(S)$ denotes the set of all sequences formed by members of $S$, allowing repetition of elements and including the empty sequence. The symbol ';' will be used to denote the sequence concatenation operator. Further, given a set $\mathcal{P}$ of propositional atoms, the set $\mathcal{L}_{\mathcal{P}}$ of literals over $\mathcal{P}$ is defined as $\mathcal{L}_{\mathcal{P}} = \mathcal{P} \cup \{\neg p \mid p \in \mathcal{P}\}$. Since no other formulae will be allowed than atoms and negated atoms, a double negation $\neg\neg p$ will be treated as identical to the unnegated atom $p$. Finally, given a set of literals $L$, the negation $Neg(L)$ of $L$ is defined as $Neg(L) = \{\neg p \mid p \in L\} \cup \{p \mid \neg p \in L\}$ and $L$ is said to be *consistent* iff there is no atom $p$ s.t. both $p \in L$ and $\neg p \in L$.

**Definition 3.2** *An instance of the* GT planning problem *is a quadruple* $\Pi = \langle \mathcal{P}, \mathcal{O}, I, G \rangle$ *where*

- $\mathcal{P}$ *is a finite set of atoms;*

- $\mathcal{O}$ *is a finite set of operators of the form* $\langle pre, post \rangle$ *where* $pre, post \subseteq \mathcal{L}_{\mathcal{P}}$ *are consistent and denote the* pre *and* post *condition respectively;*

- $I, G \subseteq \mathcal{L}_{\mathcal{P}}$ *are consistent and denote the* initial *and* goal *state respectively.*

For $o = \langle pre, post \rangle \subseteq \mathcal{O}$, we write $pre(o)$ and $post(o)$ to denote *pre* and *post* respectively. A sequence $\langle o_1, \ldots, o_n \rangle \in Seqs(\mathcal{O})$ of operators is called a *GT plan* (or simply a plan) over $\Pi$.

**Definition 3.3** *The ternary relation* valid $\subseteq Seqs(\mathcal{O}) \times 2^{\mathcal{L}_{\mathcal{P}}} \times 2^{\mathcal{L}_{\mathcal{P}}}$ *is defined s.t. for arbitrary* $\langle o_1, \ldots, o_n \rangle \in Seqs(\mathcal{O})$ *and* $S, T \subseteq \mathcal{L}_{\mathcal{P}}$, valid($\langle o_1, \ldots, o_n \rangle, S, T$) *holds iff either*

1. $n = 0$ *and* $T \subseteq S$ *or*

2. $n > 0$, $pre(o_1) \subseteq S$ and
   $valid(\langle o_2, \ldots, o_n \rangle, (S - Neg(post(o_1)) \cup post(o_1), T)$.

*A t.o. plan $\langle o_1, \ldots, o_n \rangle \in Seqs(\mathcal{O})$ solves $\Pi$ iff $valid(\langle o_1, \ldots, o_n \rangle, I, G)$.*

An action is a unique instance of an operator, *ie* a set of actions may contain several instances of the same operator, and it inherits its pre- and post-conditions from the operator it instantiates. Since all problems in this article will consider some fixed set of actions, the atom and operator sets will frequently be tacitly omitted from the GT PPIs. In figures, GT actions will be shown as boxes, with precondition literals to the left and postcondition literals to the right.

## 4. Least Constrained Plans

It seems to have been generally assumed in the planning community that there is no difference between t.o. plans and p.o. plans in the sense that a t.o. plan can easily be converted into a p.o. plan and *vice versa*. However, while a p.o. plan can be trivially converted into a t.o. plan in low-order polynomial time by topological sorting, it is less obvious that also the converse holds. At least three algorithms for converting t.o. plans into p.o. plans have been presented in the literature (Pednault, 1986; Regnier & Fade, 1991a; Veloso, Pérez, & Carbonell, 1990) (all these algorithms will be analyzed later in this article). The claim that a t.o. plan can easily be converted into a p.o. plan is vacuously true since any t.o. plan is already a p.o. plan, by definition. Hence, no computation at all needs to be done. This is hardly what the algorithms were intended to compute, however. In order to be useful, such an algorithm must output a p.o. plan satisfying some interesting criterion, ideally some optimality criterion. In fact, two of the algorithms mentioned above are claimed to produce optimal plans according to certain criteria. For instance, Veloso *et al.* (1990, p. 207) claim their algorithm to produce *least constrained* plans. They do not define what they mean by this term, however, and theirs is hardly the only paper in the literature using this term without further definition.

Unfortunately, it is by no means obvious what constitutes an intuitive or good criterion for when a p.o. plan is least constrained and, to some extent, this also depends on the purpose of achieving least-constrainment. The major motivation for producing p.o. plans instead of t.o. plans (see for instance Tate, 1975) is that a p.o. plan can be post-processed by a scheduler according to further criteria, such as release times and deadlines or resource limits. Either the actions are ordered into an (ideally) optimal sequence or, given criteria for parallel execution, into a parallel plan that can be executed faster than if the actions were executed in sequence. In both cases, the less constrained the original plan is, the greater is the chance of arriving at an optimal schedule or optimal parallel execution respectively. Both of the algorithms mentioned above are motivated by the goal of exploiting possible parallelism to decrease execution time.

It is not only interesting to make t.o. plans partially ordered, but also to make partially ordered plans more partially ordered, that is, to generalise the ordering. An algorithm for this task has been presented in the literature in the context of case-based planning (Kambhampati & Kedar, 1994). Since t.o. plan are just a special case of p.o. plans, this section will study the general problem of making partially ordered plans less constrained.

## 4.1 Least-constrainment Criteria

There is, naturally, an infinitude of possible definitions of least-constrainment. Some seem more reasonable than others, however. Three intuitively reasonable candidates are defined and analyzed below. Although other definitions are possible, it is questionable whether considerably better or more natural definitions, with respect to the purposes mentioned above, can be defined without using more information than is usually present in a t.o. or p.o. plan.

**Definition 4.1** *Let $P = \langle A, \prec \rangle$ and $Q = \langle A, \prec' \rangle$ be two p.o. plans and $\Pi$ a PPI. Then,*

1. *$Q$ is a* reordering *of $P$ wrt. $\Pi$ iff both $P$ and $Q$ are $\Pi$-valid.*

2. *$Q$ is a* deordering *of $P$ wrt. $\Pi$ iff $Q$ is a reordering of $P$ and $\prec' \subseteq \prec$*

3. *$Q$ is a* proper deordering *of $P$ wrt. $\Pi$ iff $Q$ is a reordering of $P$ and $\prec' \subset \prec$*

**Definition 4.2** *Given a PPI $\Pi$ and two p.o. plans $P = \langle A, \prec \rangle$ and $Q = \langle A, \prec' \rangle$,*

1. *$Q$ is a* minimal-constrained deordering *of $P$ wrt. $\Pi$ iff*

   (a) *$Q$ is a deordering $P$ wrt. $\Pi$ and*

   (b) *there is no proper deordering of $Q$ wrt. $\Pi$;*

2. *$Q$ is a* minimum-constrained deordering *of $P$ wrt. $\Pi$ iff*

   (a) *$Q$ is a deordering $P$ wrt. $\Pi$ and*

   (b) *there is no deordering $\langle A, \prec'' \rangle$ of $Q$ wrt. $\Pi$ s.t. $| \prec'' | < | \prec |$;*

3. *$Q$ is a* minimum-constrained reordering *of $P$ wrt. $\Pi$ iff*

   (a) *$Q$ is a reordering $P$ wrt. $\Pi$ and*

   (b) *there is no reordering $\langle A, \prec'' \rangle$ of $Q$ wrt. $\Pi$ s.t. $| \prec'' | < | \prec |$;*

Note that the previous publication (Bäckström, 1993) used the terms *LC1-minimality* for minimal-constrained deordering and *LC2-minimality* for minimum-constrained reordering. This change in terminology has been done with the hope that more will be gained in clarity than is lost by confusion.

It is easy to see that minimum-constrainment is a stronger criterion than minimal-constrainment—any minimum-constrained deordering of a plan $P$ is a minimal-constrained deordering of $P$, but the opposite is not true. As an example, consider the plan in Figure 6a. If removing all ordering constraints from action C, the result is the plan in Figure 6b, which is still valid. This plan has an order of size 3 (there is one implicit transitive order) and it is a minimal-constrained deordering since no further deordering can be made. It is not a minimum-constrained deordering, however, since if instead breaking the ordering constraints between the subsequences AB and CB, the result is the plan in Figure 6c, which is also valid. This plan has an ordering of size 2 and it can easily be seen that it is a minimum-constrained deordering, and that it happens to coincide with the minimum-constrained reordering in this case. This coincidence is not always the case, however, since a reordering is allowed to do more modifications than a deordering; a minimum deordering can obviously never have a smaller ordering relation than a minimum reordering. Examples of this difference was shown already in Section 2, where Figure 4a shows a minimum-constrained deordering and Figure 4b shows a minimum-constrained reordering.

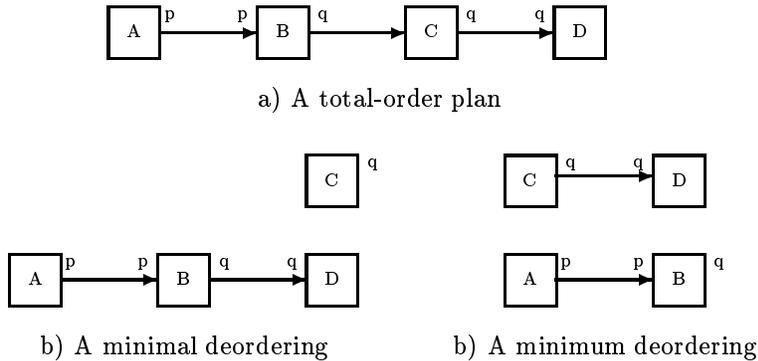

a) A total-order plan

b) A minimal deordering         b) A minimum deordering

Figure 6: The difference between minimal and minimum constrained deorderings.

Other alternative definitions of least-constrainment could be, for instance, to maximize the unorderedness or to minimize the length of the longest chain in the modified plan. However, to find a de-/reordering which has as many pairs of unordered actions as possible is the dual of computing a minimum de-/reordering and it is, thus, already covered. Minimizing the length of the longest chain is a condition which may be relevant when actions can be executed in parallel and the overall execution time is to be minimized. However, since the number of ordering constraints is quadratic in the length of a chain (because of transitive arcs), minimizing the size of the relation will often be a reasonable approximation of minimizing the chain length. Furthermore, minimizing the longest chain is still a rather weak condition for this purpose, so it is better to study directly the problem of finding shortest parallel executions of plans, which will be done later in this article.

Another issue is whether to minimize the size of the ordering relation as given, or to reduce the transitive or reductive closure of it. Since plans may have superfluous orderings with no particular purpose, it is reasonable to standardize matters and either add all possible transitive arcs, getting the transitive closure, or to remove all transitive arcs, getting the reductive closure. The choice between these two is not important for the results to be proven. However, minimizing the transitive closure will give a preference to plans with many unordered short chains of actions over plans with a few long chains, and so seems to coincide better with the term 'least constrained'.

## 4.2 Computing Least-constrained Plans

Minimal deordering is weaker than the two other least-constrainment criteria considered, but it is the least costly to achieve—it is the only one of the three criteria which can be satisfied by a polynomial-time modification to a plan.

**Definition 4.3** *The search problem* MINIMAL-CONSTRAINED DEORDERING (MLCD) *is defined as follows:*
*Given: A* PPI Π *and a* Π-*valid plan P.*
*Output: A minimal-constrained deordering of P wrt.* Π.

**Theorem 4.4** MLCD *can be solved in polynomial time if validity for p.o. plans can be tested in polynomial time.*

**Proof:** Consider algorithm MLD in Figure 7 and let $Q = \langle A, \prec' \rangle$ be the plan output by the algorithm on input $P = \langle A, \prec \rangle$. The plan $Q$ is obviously a valid deordering of $P$ wrt. Π. It is further obvious from the termination condition in the while loop that there is no other ordering $\prec'' \subset \prec'$ s.t. $\langle A, \prec'' \rangle$ is Π-valid. It follows that $Q$ is a minimal-constrained deordering. Since the algorithm obviously runs in polynomial time, the theorem follows. □

Furthermore, if validity testing is expensive, this will be the dominating cost in the MLD algorithm.

**Corollary 4.5** *If validity testing for p.o. plans can be solved in time $O(f(n))$ for some function $f(n)$, then* MLCD *can be solved in $O(\max\{n^{7/2}, n^2 f(n)\})$ time.*

```
1  procedure MLD
2     Input: A valid p.o. plan P = ⟨A, ≺⟩ and a PPI Π
3     Output: A minimal deordering of P
4  while there is some e ∈ ≺ s.t. ⟨A, (≺ −{e})⁺⟩ is Π-valid  do
5     remove e from ≺
6  return ⟨A, ≺⁺⟩;
```

Figure 7: The minimal-deordering algorithm MLD

In particular, note that plan validation is polynomial for the usual variant of propositional STRIPS without conditional actions (Nebel & Bäckström, 1994, Theorem 5.9). More precisely, this proof pertains to the *Common Propositional STRIPS formalism (CPS)* and, thus, holds also for the other common variants of propositional STRIPS, like Ground TWEAK (Bäckström, 1995). Furthermore, note that in practice it may not be necessary to compute the transitive closure either for the output plan or for validating a plan in the algorithm.

While minimum de-/reordering are stronger criteria than minimal deordering, they are also more costly to achieve.

**Definition 4.6** *The decision problem* MINIMUM-CONSTRAINED DEORDERING (MMCD) *is defined as follows:*
*Given: A* PPI Π, *a* Π-*valid plan P and an integer $k \geq 0$.*
*Question: Is there a deordering $\langle A, \prec \rangle$ of P s.t. $| \prec | \leq k$?*

**Definition 4.7** *The decision problem* MINIMUM-CONSTRAINED REORDERING (MMCR) *is defined as follows:*
*Given: A* PPI $\Pi$, *a* $\Pi$-*valid plan* $P$ *and an integer* $k \geq 0$.
*Question: Is there a reordering* $\langle A, \prec \rangle$ *of* $P$ *s.t.* $| \prec | \leq k$?

**Theorem 4.8** MINIMUM-CONSTRAINED DEORDERING *is NP-hard.*

**Proof:** Proof by reduction from MINIMUM COVER (Garey & Johnson, 1979, p. 222), which is NP-complete. Let $S = \{p_1, \ldots, p_n\}$ be a set of atoms, $C = \{C_1, \ldots, C_m\}$ a set of subsets of $S$ and $k \leq |C|$ a positive integer. A *cover* of size $k$ for $S$ is a subset $C' \subseteq C$ s.t. $|C'| \leq k$ and $S \subseteq \cup_{T \in C'} T$. Construct, in polynomial time, the GT PPI $\Pi = \langle \emptyset, \{r\} \rangle$ and the $\Pi$-valid t.o. plan $P = \langle a_1, \ldots, a_m, a_S \rangle$ where $pre(a_i) = \emptyset$ and $post(a_i) = C_i$ for $1 \leq i \leq m$, and further $pre(a_S) = S$ and $post(a_S) = \{r\}$. Obviously, $S$ has a minimum cover of size $k$ iff there exists some $\Pi$-valid p.o. plan $Q = \langle \{a_1, \ldots, a_m, a_S\}, \prec \rangle$ s.t. $| \prec | \leq k$, since only those actions contributing to the cover need remain ordered wrt. to $a_S$ $\qquad \square$

**Corollary 4.9** MINIMUM-CONSTRAINED REORDERING *is NP-hard.*

**Corollary 4.10** MINIMUM-CONSTRAINED DEORDERING *and* MINIMUM-CONSTRAINED REORDERING *both remain NP-hard even when restricted to GT plans where the actions have only positive pre- and post-conditions.*

**Theorem 4.11** *If validity for p.o. plans is in some complexity class* $C$, *then* MINIMUM-CONSTRAINED DEORDERING *and* MINIMUM-CONSTRAINED REORDERING *are in* $NP^C$.

**Proof:** Guess a solution, verify that it is a de-/reordering and then validate it using an oracle for C. $\qquad \square$

For most common planning formalisms without conditional actions and context-dependent effects, minimal de-/reordering is NP-complete.

**Theorem 4.12** *If validity for p.o. plans can be tested in polynomial time, then* MINIMUM-CONSTRAINED DEORDERING *and* MINIMUM-CONSTRAINED REORDERING *are NP-complete.*

**Proof:** Immediate from Theorems 4.8 and 4.11 and from Corollary 4.9. $\qquad \square$

It follows immediately that the corresponding search problems, that is, the problems of *generating* a minimum-constrained de-/reordering are also NP-hard (and even NP-equivalent if validity testing is tractable).

Furthermore, MMCD and MMCR are not only hard to solve optimally, but even to approximate. Neither of these problems is in the approximation class APX (Crescenzi & Panconesi, 1991), *ie* neither problem can be approximated within a constant factor. (Both here and elsewhere in this article the term approximation is used in the constructive sense, that is the results refer to the existence/non-existence of algorithms producing an approximate solution in polynomial time).

**Theorem 4.13** MINIMUM-CONSTRAINED DEORDERING *and* MINIMUM- CONSTRAINED REORDERING *cannot be approximated within a constant unless* $NP \in$ DTIME$(n^{poly\ log\ n})$.

**Proof:** Suppose there were a polynomial-time algorithm A approximating MMCD within a constant. Since the reduction in the proof of Theorem 4.8 preserves the solutions exactly, also approximations are preserved. Hence, MINIMUM COVER could be approximated within a constant, but this is impossible unless $NP \in$ DTIME$(n^{poly\ log\ n})$ (Lund & Yannakakis, 1994), which contradicts the assumption. The case for MMCR is a trivial consequence. $\square$

If using the number of propositional atoms in the plan as a measure of its size, this bound can be strengthened to $(1 - \varepsilon) \ln |\mathcal{P}|$ for arbitrary $\varepsilon$ unless $NP \in$ DTIME$(n^{\log \log\ n})$ by substituting such a result for MINIMUM COVER (Feige, 1996) in the proof above.

## 5. Parallel Plans

In order to study the problem of finding a shortest parallel execution of a plan, the formalisms used so far are not quite sufficient. Since they lack a capability of modelling when actions can be executed in parallel or not, it is impossible to say with any reasonable precision how a certain action ordering will affect the parallel execution time. Partial-order plans are sometimes referred to as parallel plans in literature. This is misleading, however. That two actions are left unordered in such a plan means that they can be executed in either order, without affecting the validity of the plan, but in the general case there is no guarantee that the plan will remain valid also if the executions of the actions overlap temporally. In some cases, unorderedness means that parallel or overlapping execution is allowed, while in other cases it does not mean that, depending on the action modelling and its underlying domain assumptions. In the first case, the plan must have a stronger ordering committment, any two actions that must not have overlapping executions must be ordered, thus making the plan over-committed.

In order to distinguish the two cases, a concept of parallel plans will be introduced below. A parallel plan is a partial-order plan with an extra relation, a *non-concurrency relation*, which tells which actions must not be executed in parallel. In this article two actions are considered parallel if their executions have any temporal overlap at all. Plans where all unordered actions can be executed in parallel constitute the special case of *definite* parallel plans.

**Definition 5.1** *A* parallel plan *is a triple* $P = \langle A, \prec, \# \rangle$*, where* $\langle A, \prec \rangle$ *is a p.o. plan and* $\#$ *is an irreflexive, symmetric relation on* $A$*. A* definite parallel p.o plan *is a parallel plan* $P = \langle A, \prec, \# \rangle$ *s.t.* $\# \subseteq (\prec \cup \prec^{-1})$.

Intuitively, a parallel plan is a p.o. plan extended with an extra relation, $\#$ (a *non-concurrency* relation), expressing which of the actions must not be executed in parallel. This relation is primarily intended to convey information about actions that are unordered under the $\prec$ relation, although it is allowed to relate also such actions. That is, the $\#$ relation is intended to capture information about whether two actions can be executed in parallel or not, in general. That two actions are ordered in a plan forbids executing them in parallel in this particular plan, but does not necessarily mean that the actions could not be executed in parallel under different circumstances. Planning algorithms frequently produce overcommitted orderings on plans, and the whole purpose of this article is to study the problem of optimizing plans by finding and removing such overcommitted orderings. Hence, there are no restrictions in general on the relation $\#$ in addition to those in Definition 5.1. For instance, $a \prec b$ does not imply that $a\#b$. However, the non-concurrency relation will frequently be constrained to satisfy the post-exclusion principle.

**Definition 5.2** *A parallel GT plan* $P = \langle A, \prec, \# \rangle$ *satisfies the* post-exclusion principle *iff for all actions* $a, b \in A$, $a\#b$ *whenever there is some atom* $p$ *s.t.* $p \in post(a)$ *and* $\neg p \in post(b)$.

The definition of plan validity is directly inherited from p.o. plans.

**Definition 5.3** *Given a* PPI $\Pi$, *a parallel plan* $\langle A, \prec, \# \rangle$ *is* $\Pi$-*valid iff the p.o. plan* $\langle A, \prec \rangle$ *is* $\Pi$-*valid.*

The non-concurrency relation is, thus, not relevant for deciding whether a plan is valid or not. Instead, it is used for constraining how parallel plans may be executed and it is the core concept behind the definition of parallel executions.

Consider, for instance, the GT plan $\langle \{A, B, C\}, \{\langle A, B \rangle\}, \{\langle B, C \rangle\} \rangle$ which is shown in Figure 8 (arrows denote ordering relations and dashed lines denote nonconcurrency relations). This plan is valid wrt. the PPI $\Pi = \langle \emptyset, \{r, s\} \rangle$, that is the final value of the atom $q$ does not matter. Since $B\#C$ holds the actions B and C are constrained not to be executed in parallel, but may be executed in either order, that is, the plan is not definite. This could be because the post-exclusion principle is employed, or for some other reason. Although $A\#B$ does not hold the actions A and B clearly cannot be executed in parallel, since $A \prec B$ holds. There are four ways to execute this plan, in either of the three sequences A,B,C; A,C,B and C,A,B, or by executing A and C in parallel, followed by B (unit length is assumed). Also note that this plan would no longer be valid if the goal contained either $q$ or $\neg q$, since the final truth value of $q$ depends on the actual execution order. Furthermore, any reordering of the plan would have to keep the ordering constraint $A \prec B$ to satisfy the validity criterion, why it is not necessary to have the constraint $A\#B$. It would do no harm here to include this restriction, but in more complex plans it may be an over-constrainment, if there are several producers for the atom $p$ to choose between, for instance. To sum up, the non-concurrency relation should primarily be used to mark which actions must not be in parallel in addition to those already forbidden to be in parallel because of validity.

This framework for parallel plans admits expressing possible parallelism only; necessary parallelism is out of the scope of this article and requires a planner having access to and being able to make use of further additional information, perhaps a temporal algebra. Furthermore, a set of non-concurrent actions can easily be expressed by making all actions in the set pairwise non-concurrent, but the formalism is not sufficient to say that $k$ of the actions, but not more, in such a set may be executed in parallel. Similarly, it is not possible to express that an action must executed before or after an interval, or that two sets of actions must have non-overlapping executions.

**Definition 5.4** *Let* $P = \langle A, \prec, \# \rangle$ *be a parallel plan and let the function* $d : A \mapsto \mathbb{N}$ *denote the duration of each action. A* parallel execution *of* $P$ *is a function* $r : A \mapsto \mathbb{N}$, *denoting release times for the actions in* $A$, *satisfying that for all* $a, b \in A$,

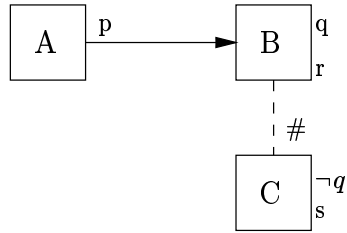

Figure 8: A parallel plan

1. *if* $a \prec b$, *then* $r(a) + d(a) \leq r(b)$ *and*

2. *if* $a \# b$, *then either*

   (a) $r(a) + d(a) \leq r(b)$ *or*

   (b) $r(b) + d(b) \leq r(a)$.

*The* length *of the parallel execution is defined as* $\max_{a \in A}\{r(a) + d(a)\}$, *ie, the latest finishing time of any action. A* minimum parallel execution *of plan is a parallel execution with minimum length among all parallel executions of the plan. The* length *of a parallel plan* $P$, *denoted length*$(P)$, *is the length of the minimum parallel execution(s) for* $P$.

Obviously, every parallel plan has a parallel execution of length $\sum_{a \in A} d(a)$ (which is the trivial case of sequential execution). Furthermore, in certain cases, hardness results will be strengthened by restricting the duration function.

**Definition 5.5** *The special case where* $d(a) = 1$ *for all* $a \in A$ *is referred to as the* unit time assumption.

Deciding whether a release-time function is a parallel execution is tractable.

**Theorem 5.6** *Given a parallel plan* $P = \langle A, \prec, \# \rangle$, *a duration function* $d : A \mapsto \mathbb{N}$ *and a release-time function* $r : A \mapsto \mathbb{N}$, *it can be decided in polynomial time whether* $r$ *is a parallel execution for* $P$ *and, in the case it is, what the length of this execution is.*

**Proof:** Trivial. □

Consider the plan in Figure 8 and three release-time functions $r_1$, $r_2$ and $r_3$, defined as follows

$$r_1(A) = 1 \quad r_1(B) = 2 \quad r_1(C) = 3$$
$$r_2(A) = 1 \quad r_2(B) = 2 \quad r_2(C) = 1$$
$$r_3(A) = 1 \quad r_3(B) = 2 \quad r_3(C) = 2.$$

Both $r_1$ and $r_2$ are parallel executions of the plan, while $r_3$ is not. Furthermore, $r_2$ is a minimum parallel execution for the plan, having length 2. However, computing the minimum parallel execution of a parallel plan is difficult in the general case.

**Definition 5.7** *The decision problem* PARALLEL PLAN LENGTH (PPL) *is defined as follows:*
*Given: A parallel plan $P = \langle A, \prec, \# \rangle$, a duration function $d$ and an integer $k$.*
*Question: Does $P$ have a parallel execution of length $k$ or shorter?*

**Theorem 5.8** PARALLEL PLAN LENGTH *is NP-hard.*

**Proof:**   Hardness is proven by transformation from GRAPH K-COLOURABILITY (Garey & Johnson, 1979, p. 191), which is NP-complete. Let $G = \langle V, E \rangle$ be an arbitrary undirected graph, where $V = \{v_1, \ldots, v_n\}$. Construct, in polynomial time, a GT PPI as follows. Define the PPI $\Pi = \langle \emptyset, \{p_1, \ldots, p_n\} \rangle$. Also define the parallel plan $P = \langle A, \emptyset, \# \rangle$, where $A$ contains one action $a_i$ for each vertex $v_i \in V$, s.t. $pre(a_i) = \emptyset$ and $post(a_i) = \{p_i, q_i\} \cup \{\neg q_j \mid \{v_i, v_j\} \in E\}$. Finally, let $a_i \# a_j$ iff $\{v_i, v_j\} \in E$, which satisfies the post-exclusion principle. The plan $P$ just constructed is obviously $\Pi$-valid. It is easy to see that $G$ is $k$-colourable iff $P$ has a parallel execution of length $k$ wrt. $\Pi$ since each colour of $G$ will correspond to a unique release time in the parallel execution of $P$.   □

**Corollary 5.9** PARALLEL PLAN LENGTH *remains NP-hard even when restricted to GT actions with empty preconditions and under the assumption of unit time and the post-exclusion principle.*

**Theorem 5.10** PARALLEL PLAN LENGTH *is in NP.*

**Proof:**   Guess a parallel execution. Then verify it, which can be done in polynomial time according to Theorem 5.6.   □

Computing a minimum parallel execution of a plan is tractable for the special case of definite plans, however.

**Theorem 5.11** PARALLEL PLAN LENGTH *can be solved in polynomial time for definite parallel plans.*

**Proof:**   Use the algorithm DPPL (Figure 9), which is a straightforward stratification algorithm for directed DAGs.   □

## 6. Reordering Parallel Plans

Having defined the concept of parallel plan, it is possible to define concepts similar to the previous least-constrainment criteria which are more appropriate for minimizing the execution time of parallel plans.

**Definition 6.1** *Let $P = \langle A, \prec, \# \rangle$ and $Q = \langle A, \prec', \# \rangle$ be two parallel plans and $\Pi$ a PPI. Then,*

*1. $Q$ is a* parallel reordering *of $P$ wrt. $\Pi$ iff both $P$ and $Q$ are $\Pi$-valid;*

```
1   procedure DPPL
2       Input: A definite parallel plan P = ⟨A, ≺, #⟩
3       Output: A minimum parallel execution r for P
4   Construct the directed graph G = ⟨A, ≺⟩
5   for all a ∈ A  do
6       r(a) ← 0
7   while A ≠ ∅  do
8       Select some node a ∈ A without predecessors in A
9       for all b ∈ A s.t.  a ≺ b  do
10          r(b) ← max(r(b), r(a) + d(a))
11      A ← A − {a}
12  return r
```

Figure 9: Algorithm for computing a minimum parallel execution for definite parallel plans.

2. $Q$ is a parallel deordering *of $P$ wrt.* $\Pi$ *iff $Q$ is a parallel reordering of $P$ and $\prec' \subseteq \prec$;*

3. $Q$ is a minimum parallel reordering *of $P$ wrt.* $\Pi$ *iff*

   (a) *$Q$ is a parallel reordering of $P$ wrt.* $\Pi$ *and*

   (b) *no other parallel reordering of $P$ wrt.* $\Pi$ *is of shorter length than $Q$;*

4. $Q$ is a minimum parallel deordering *of $P$ wrt.* $\Pi$ *iff*

   (a) *$Q$ is a parallel deordering of $P$ wrt.* $\Pi$ *and*

   (b) *no other parallel deordering of $P$ wrt.* $\Pi$ *is of shorter length than $Q$.*

Modifying plans to satisfy either of the latter two criteria is difficult in the general case, however.

**Definition 6.2** *The decision problem* Minimum Parallel Deordering (MmPD) *is defined as follows.*
*Given: a* PPI $\Pi$*, a parallel plan $P$, a duration function $d$ and an integer $k$.*
*Question: Does $P$ have a deordering with a parallel execution of length $k$ wrt.* $\Pi$*?*

**Definition 6.3** *The decision problem* Minimum Parallel Reordering (MmPR) *is defined as follows.*
*Given: a* PPI $\Pi$*, a parallel plan $P$, a duration function $d$ and an integer $k$.*
*Question: Does $P$ have a reordering with a parallel execution of length $k$ wrt.* $\Pi$*?*

**Theorem 6.4** Minimum Parallel Deordering *is NP-hard.*

**Proof:**   Similar to the proof of Theorem 6.4. Given a graph $G$ and an integer $k$, construct a PPI $\Pi$ and a plan $P = \langle A, \prec, \# \rangle$ in the same way as in the proof of Theorem 5.8, but let $\prec$ be an arbitrary total order on $A$. Obviously, $P$ is $\Pi$-valid and $Q = \langle A, \emptyset, \# \rangle$ is a deordering of $P$ s.t. no other deordering of $P$ is shorter than $Q$. Hence, $Q$, and thus $P$, has a deordering with a parallel execution of length $k$ iff $G$ is $k$-colourable.   □

**Corollary 6.5** MINIMUM PARALLEL REORDERING *is NP-hard.*

**Corollary 6.6** MINIMUM PARALLEL DEORDERING *and* MINIMUM PARALLEL REORDER-ING *remain NP-hard even when restricted to totally ordered GT plans and under the assumptions of unit time and simple concurrency.*

Note that the restriction to definite input plans is covered by this corollary. If output plans are also required to be definite, then the reordering case remains NP-hard.

**Theorem 6.7** MINIMUM PARALLEL REORDERING *remains NP-hard also when the output plan is restricted to be definite.*

**Proof:** Reuse the proof for Theorem 6.4 as follows. Let $r$ be a shortest parallel execution for the plan $Q$ and assume this execution is of length $n$. Construct an order $\prec'$ on $A$ s.t. for all actions $a, b \in A$, $a \prec' b$ iff $r(a) < r(b)$. Obviously the plan $\langle A, \prec', \# \rangle$ is a definite minimum parallel reordering of $P$. It follows that $P$ has a definite parallel reordering of length $k$ iff $G$ is $k$-colourable. □

It is an open question whether minimum deordering remains NP-hard when also output plans must be definite, but an important special case is polynomial, as will be proven in the next section.

**Theorem 6.8** MINIMUM PARALLEL DEORDERING *and* MINIMUM PARALLEL REORDER-ING *are in* $NP^C$ *if validation of p.o. plans is in some complexity class $C$.*

**Proof:** Given a plan $\langle A, \prec, \# \rangle$, a duration function $d$ and a parameter $k$, guess a de/reordering $\prec'$ and a release-time function $r$. Then verify, using an oracle for $C$, that $\langle A, \prec', \# \rangle$ is valid. Finally, verify that $r$ is a parallel execution of length $\leq k$, which is polynomial according to Theorem 5.6. □

**Theorem 6.9** *Minimum parallel de-/reordering is NP-complete if p.o. plans can be validated in polynomial time.*

**Proof:** Immediate from Theorems 6.4 and 6.8 and Corollary 6.5. □

The problems MMPD and MMPR are not only hard to solve optimally, but also to approximate.

**Theorem 6.10** MINIMUM PARALLEL DEORDERING *and* MINIMUM PARALLEL REORDER-ING *cannot be approximated within* $|A|^{1/7-\varepsilon}$ *for any $\varepsilon > 0$, unless P=NP.*

**Proof:** Suppose there were a polynomial-time algorithm A approximating MMCD within $|A|^{1/7-\varepsilon}$ for some $\varepsilon > 0$. Then it is immediate from the proof of Theorem 6.4 that also GRAPH K-COLOURABILITY could be approximated within $|A|^{1/7-\varepsilon}$, which is impossible unless P=NP (Bellare, Goldreich, & Sudan, 1995). □

With the same reasoning, this bound can be strengthened to $|A|^{1-\varepsilon}$, under the assumption that co-RP≠NP (Feige & Kilian, 1996).

## 7. Restricted Cases

Since the problems of computing minimum de-/reorderings are very difficult, and are even difficult to approximate, an alternative way of tackling them could be to study restricted cases. One special case already considered is the restriction to definite plans only. While the problem MMPR is still NP-complete under this restriction, it is an open question whether also MMPD is NP-complete. A positive result can be proven, though, to the effect that MMPD is polynomial for definite plans for a large class of planning languages, including most of the commonly used ones. This result will be proven by generalising an algorithm from the literature for deordering total-order plans.

Based on the (not necessarily true) argument that it is easier to generate a t.o. plan than a p.o. plan when using complex action representations, Regnier and Fade (1991a, 1991b) have presented an algorithm for converting a t.o. plan into a p.o. plan. The resulting plan has the property that all its unordered actions can be executed in parallel, that is, the plan is definite. The authors of the algorithm further claim that the algorithm finds all pairs of actions that can be executed in parallel and, hence, the plan can be post-processed to find an optimal parallel execution. They do not define what they mean by this criterion, however.

Incidentally, the algorithm proposed by Regnier and Fade is a special case of an algorithm earlier proposed for the same problem by Pednault (1986), who did not make any claims about optimality. If removing from Regnier and Fade's algorithm all details relevant only for their particular implementation and planning language, the two algorithms coincide and they are thus presented here as one single algorithm, the PRF algorithm[4] (Figure 10). PRF is slightly modified from the original algorithms. First, it does not assume that the input plan is totally ordered, since it turns out to be sufficient that it is a definite partial-order plan. Second, PRF returns a parallel plan, rather than a p.o. plan—a harmless modification since the only additional piece of information is the non-concurrency relation, which is already given as input, either explicitly or implicitly. Third, PRF returns the transitive closure of its ordering relation. This is by no means necessary, and is motivated, as usual, by conforming to the definitions of this article.

1    **procedure** PRF;
2       Input: A PPI $\Pi$, a $\Pi$-valid definite p.o. plan $\langle A, \prec \rangle$ and a non-concurrency relation $\#$
3       Output: A $\Pi$-valid parallel plan
4    **for** all $a, b \in A$ s.t. $a \prec b$ **do**
5       **if** $a \# b$ **then**
6          Order $a \prec' b$;
7    **return** $\langle A, \prec'^+, \# \rangle$;

Figure 10: The PRF algorithm

Obviously, PRF computes a deordering of its input, and it is unclear whether it is possible to compute a minimal definite deordering in polynomial time. However, the algorithm has been abstracted here to a very general formalism, and an analysis for restricted formalisms reveals more about its performance. The language used by Regnier and Fade is unnecessarily restricted so the algorithm will be shown to work for a considerably more general formalism, based on generalising and abstracting the concepts of producers, consumers and threats used in most common planners and planning languages, *eg* STRIPS and TWEAK. This formalism will be referred to as the *Producer-Consumer-Threat formalism (PCT)*.

---

4. Here and afterwards, the algorithms from the literature will be referred to by acronyms consisting of the initials of its authors, in this case Pednault, Regnier and Fade.

Let $prod(a, \pi)$ denote that $a$ produces the condition $\pi$, $cons(a, \pi)$ that $a$ consumes $\pi$ and $threat(a, \pi)$ that $a$ is a threat to $\pi$. To simplify the definitions, the standard transformation will be used of simulating the initial and goal states with actions. That is, every PCT plan contains an action ordered before all other actions which consumes nothing and produces the initial state. Similarly, there is an action ordered after all other actions which consumes the goal state and produces nothing. This means that the PPI is contained within the plan itself, so all references to PPIs can be omitted in the following. Validity of plans can then be defined as follows.

**Definition 7.1** *A t.o. PCT plan $\langle a_1, \ldots, a_n \rangle$ is valid iff for all $i$, $1 \leq i \leq n$ and all conditions $\pi$ s.t. $cons(a_i, \pi)$, there is some $j$, $1 \leq j < i$ s.t. $prod(a_j, \pi)$ and there is no $k$, $j \leq k \leq i$ s.t. $threat(a_k, \pi)$. A p.o. PCT plan is valid iff all topological sortings of it are valid.*

Chapman's Modal-truth Criterion (MTC) (Chapman, 1987) can be abstracted to the PCT formalism and be analogously used for validating p.o. plans.

**Definition 7.2** *The* modal truth criterion (MTC) *for a PCT plan $\langle A, \prec \rangle$ is:*

$$\forall a_C \forall \pi (cons(a_C, \pi) \rightarrow$$
$$\exists a_P (prod(a_P, \pi) \wedge a_P \prec a_C \wedge$$
$$\forall a_T (threat(a_T, \pi) \rightarrow$$
$$a_C \prec a_T \vee$$
$$\exists a_W (prod(a_W, \pi) \wedge a_T \prec a_W \wedge a_W \prec a_C))))$$

**Theorem 7.3** *The MTC holds for a PCT plan P iff it is valid.*

**Proof:** Trivial generalization of the proofs leading to Theorem 5.9 in Nebel and Bäckström (1994). □

Only a minimum of constraints for when two actions may not be executed in parallel will be required. These constraints are obeyed by most planners in the AI literature.

**Definition 7.4** Simple concurrency *holds if for all actions $a$, $b$ s.t. $a \neq b$, the non-concurrency relation satisfies the following three conditions*

1. $prod(a, \pi) \wedge cons(b, \pi) \rightarrow a \# b$

2. $prod(a, \pi) \wedge threat(b, \pi) \rightarrow a \# b$

3. $cons(a, \pi) \wedge threat(b, \pi) \rightarrow a \# b$

Note that it is not required that two producers, two consumers or two threats of the same condition are non-concurrent, thus allowing, for instance, plans with multiple producers, *eg* Nebel and Bäckström (1994, Fig. 4) and Kambhampati (1994). The axioms do not prevent adding such restrictions, though. Furthermore, note that the definition only states a necessary condition for non-concurrency—it is perfectly legal to add further non-concurrency constraints on the actions in a plan. It may also be worth noting that the MTC requires producers and threats to be ordered only if there is a corresponding consumer, while a definite plan satisfying the simple concurrency criterion always require them to be ordered.

The following observation about PRF is immediate from the algorithm and will be used in the proofs below.

**Observation 7.5** *If $\langle A, \prec, \# \rangle$ is the input to PRF and $\langle A, \prec', \# \rangle$ is the corresponding output, then it holds that $a \prec' b$ iff $a \prec b$ and $a \# b$.*

Based on this lemma, it can be proven that PRF preserves validity.

**Lemma 7.6** *If the plan input to PRF is a valid PCT plan and $\#$ satisfies the simple concurrency criterion, then the output plan is valid.*

**Proof:** Let $P = \langle A, \prec, \# \rangle$ be the input plan and $Q = \langle A, \prec', \# \rangle$ the output plan. Since $P$ is valid, it follows from Theorem 7.3 that the MTC holds for $P$. Adding the implied simple-concurrency constraints to the MTC yields the following condition:

$$\forall a_C \forall \pi (cons(a_C, \pi) \rightarrow$$
$$\exists a_P (prod(a_P, \pi) \wedge a_P \prec a_C \wedge a_P \# a_C \wedge$$
$$\forall a_T (threat(a_T, \pi) \rightarrow$$
$$(a_C \prec a_T \wedge a_C \# a_T) \vee$$
$$\exists a_W (prod(a_W, \pi) \wedge a_T \prec a_W \wedge a_T \# a_W \wedge$$
$$a_W \prec a_C \wedge a_W \# a_C)))).$$

By applying Observation 7.5 this can be simplified to:

$$\forall a_C \forall \pi (cons(a_C, \pi) \rightarrow$$
$$\exists a_P (prod(a_P, \pi) \wedge a_P \prec' a_C \wedge$$
$$\forall a_T (threat(a_T, \pi) \rightarrow$$
$$a_C \prec' a_T \vee$$
$$\exists a_W (prod(a_W, \pi) \wedge a_T \prec' a_W \wedge a_W \prec' a_C)))),$$

which is the MTC for the plan $Q$. Once again using Theorem 7.3, it follows that $Q$ is valid. $\square$

This allows for proving that PRF produces definite minimum deorderings of definite PCT plans under simple concurrency.

**Theorem 7.7** *If using the PCT formalism and simple concurrency, then PRF produces a minimum-deordered definite version of its input.*

**Proof:** Let $P = \langle A, \prec, \# \rangle$ be the input plan, which is assumed valid and definite, and $Q = \langle A, \prec', \# \rangle$ the output plan. It is obvious that $\prec' \subseteq \prec$ and it follows from Lemma 7.6 that $Q$ is valid, so $Q$ is a deordering of $P$. It remains to prove that $Q$ is a minimum deordering of $P$.

Suppose that $P$ has a deordering $R = \langle A, \prec'', \# \rangle$ s.t. $|\prec''| < |\prec'|$. Then, there must be some $a, b \in A$ s.t. $a \prec' b$, but not $a \prec'' b$. It can be assumed that $a \prec' b$ is not a transitive arc in $\prec'$, since the transitive closure is anyway computed at the end of the algorithm. Since the order $\prec'$ is produced by PRF, it follows from Observation 7.5 that $a \prec b$ and $a \# b$. Because of the latter constraint, it is necessary that either, $a \prec'' b$ or $b \prec'' a$ holds, but only the former is possible since $a \prec b$ and $R$ is a deordering of $P$. This contradicts the assumption, so $Q$ must be a minimum deordering of $P$. □

Since PRF is a polynomial algorithm, it follows that definite minimum deorderings of definite PCT plans can be computed in polynomial time under simple concurrency. Furthermore, since PRF produces definite plans it is possible to actually compute the shortest parallel execution efficiently.

**Theorem 7.8** *If the plan input to PRF is a valid and definite PCT plan satisfying the simple concurrency criterion, then PRF outputs a definite minimum deordering of this plan.*

**Proof:** PRF runs in polynomial time and obviously produces definite parallel plans. Hence, it follows from Theorem 5.11 that a minimum parallel execution for the output plan can be found in polynomial time, which proves the theorem. □

It seems likely that this is what Regnier and Fade meant with their optimality claim, although for a special instance of the PCT formalism. This result says nothing about the difficulty of finding a minimum reordering of a plan, since PRF only considers deorderings. Since minimum deorderings do not approximate minimum reorderings well, it can be suspected that it is more difficult to compute the latter. The following theorem confirms this suspicion, showing that the latter problem remains NP-hard under quite severe restrictions, including the following two.

**Definition 7.9** *A GT action $a$ is* toggling *iff for all literals $l \in post(a)$, it is also the case that $\neg l \in pre(a)$. A GT action $a$ is* unary *iff $|post(a)| = 1$.*

**Theorem 7.10** MINIMUM PARALLEL REORDERING *remains NP-hard even when restricted to total-order GT plans with only toggling unary actions and under the assumption of unit time, simple concurrency and that no actions are redundant.*

The proof of this theorem appears in Appendix A.

While minimum reorderings are more difficult to compute than minimum deorderings, they can also produce arbitrarily better results.

**Theorem 7.11** MINIMUM PARALLEL DEORDERING *cannot approximate* MINIMUM PARALLEL REORDERING *within $|A|^k$ for any constant $k \geq 0$.*

The proof of this theorem appears in Appendix A.

**Corollary 7.12** MINIMUM PARALLEL DEORDERING *cannot approximate* MINIMUM PARALLEL REORDERING *within* $|A|^k$ *for any constant* $k \geq 0$ *even when the problems are restricted to GT plans with only positive preconditions and under the assumption of simple concurrency.*

It may, thus, appear as though minimum reordering is a preferable, albeit more costly, operation than minimum deordering. However, if the plan modification is to be followed by scheduling, it is no longer obvious that a reordering is to prefer. Since scheduling may take further information and constraints into account, *eg* upper and lower bounds on the release time and limited resources, a feasible schedule for the original plan may no longer be a feasible schedule for a reordering of the same plan. That is, some or all feasible solutions may be lost when reordering a plan. In contrast to this, deordering a plan is harmless since all previously feasible schedules are preserved in the deordering. Of course, the de-/reordered plan may have new and better schedules than the old plan, which is why the problems studied in this article are interesting at all. However, while minimum deordering is a safe and, usually cheap, operation, minimum reordering is neither and must thus be applied with more care. To find a reordering of a plan with an optimum schedule would require combining minimum reordering and scheduling into one single computation, but it is out of the scope of this article to study such combinations. Suffice it to observe that such a computation is never cheaper than either of its constituent computations.

## 8. Related work

This section analyses and discusses some algorithms suggested in the literature for generalising the ordering of a plan, in addition to the PRF algorithm already analysed in the preceeding section. Also some planners that generate plans with some optimality flavour on the ordering are discussed.

Some of the algorithms to be analysed use the common trick of simulating the initial state and the goal of a planning instance by two extra operators, in the following way. Let $P = \langle A, \prec \rangle$ be a plan and $\Pi = \langle I, G \rangle$ a PPI, both in the GT language. Introduce two extra actions $a_I$, with $pre(a_I) = \emptyset$ and $post(a_I) = I$, and $a_G$, with $pre(a_G) = G$ and $post(a_G) = \emptyset$. Define the plan $Q = \langle A \cup \{a_I, a_G\}, \prec' \rangle$ where $\prec' = \prec \cup \{a_I \prec a, a \prec a_G \mid a \in A\} \cup \{a_I \prec a_G\}$, that is $a_I$ is ordered before all other actions and $a_G$ is ordered after all other actions. The plan $Q$ is a representation of both the plan $P$ and the PPI $\Pi$. Such a combined representation will be referred to as a *self-contained plan*. A self-contained plan is valid iff it is valid wrt. to the PPI $\langle \emptyset, \emptyset \rangle$. It is trivial to convert a plan and a PPI into a corresponding self-contained plan and vice versa. Hence, both ways of representing a plan will be used alternately without further notice.

### 8.1 The VPC Algorithm

Veloso *et al.* (1990) have presented an algorithm (here referred to as VPC[5]) for converting t.o. plans into 'least-constrained' p.o. plans. They use the algorithm in the following context. First a total-order planner (NOLIMIT) is used to produce a t.o. plan. VPC converts this plan

---

5. In the original publication the algorithm was named Build_Partial_Order.

```
1   procedure VPC;
2      Input: a valid self-contained t.o. plan ⟨a₁, . . . , aₙ⟩
              where a₁ = a_I and aₙ = a_G
3      Output: A self-contained valid p.o. plan
4   for 1 ≤ i ≤ n  do
5      for p ∈ pre(aᵢ)  do
6         Find max k < i s.t.  p ∈ post(a_k);
7         if such a k exists  then
8            Order a_k ≺ aᵢ
9      for ¬p ∈ post(aᵢ)  do
10        for 1 ≤ k < i s.t.  p ∈ pre(a_k)  do
11           Order a_k ≺ aᵢ
12     for each primary effect p ∈ post(aᵢ)  do
13        for 1 ≤ k ≤ i s.t.  ¬p ∈ post(a_k)  do
14           Order aᵢ ≺ a_k
15     for 1 < i < n  do
16        Order a_I ≺ aᵢ and aᵢ ≺ a_G
17     return ⟨{a₁, . . . , aₙ}, ≺⁺⟩;
```

Figure 11: The VPC algorithm

into a p.o. plan which is then post-processed to determine which actions can be executed in parallel. The action language used is a STRIPS-style language allowing quantifiers and context-dependent effects. However, the plans produced by the planner, and thus input to VPC, are ground and without context-dependent effects. That is, they are ordinary propositional STRIPS plans. The VPC algorithm is presented in Figure 11, with a few minor differences in presentation as compared to its original appearance: First, the algorithm is presented in the GT formalism, in order to minimize the number of formalisms in this article, but all preconditions are assumed to be positive, thus coinciding with the original algorithm. Second, while the original algorithm returns the transitive reduction of the computed order it instead returns the transitive closure here, an unimportant difference in order to coincide with the definition of plans in this article. Furthermore, Veloso[6] has pointed out that the published version of the VPC algorithm is incorrect and that a corrected version exists. The version presented in Figure 11 is this corrected version. A proposition is a primary effect if it appears either in the goal or in the subgoaling chain of a goal proposition.

VPC is a greedy algorithm which constructs an entirely new partial order by analysing the action conditions, using the original total order only to guide the greedy strategy. The algorithm is claimed (Veloso et al., 1990, p. 207) to produce a 'least-constrained' p.o. plan, although no definition is given of what this means. Veloso[7] has confirmed that the term 'least constrained plan' was used in a 'loose sense' and no optimality claim was intended. However, if this term is not defined, then it is impossible to know what problem the algorithm is intended to solve or how to judge whether it makes any improvement over using no algorithm at all. In the absence of such a definition from its authors, the algorithm will be analysed with respect to the least-constraint criteria defined in Section 4. This is admittedly a somewhat unfair analysis, but it reveals some interesting facts about the algorithm, and about what problems it does not solve. It is immediate from Theorem 4.8 and Corollary 4.9 that VPC cannot be expected to produce minimum-constrained de-/reorderings. Perhaps more surprisingly, VPC does not even guarantee that its output is a *minimal*-constrained deordering of its input, a problem already proven trivially polynomial (Theorem 4.4). This is illustrated by the following example.

---

6. Personal communication, oct. 1993.
7. Veloso, *ibid.*

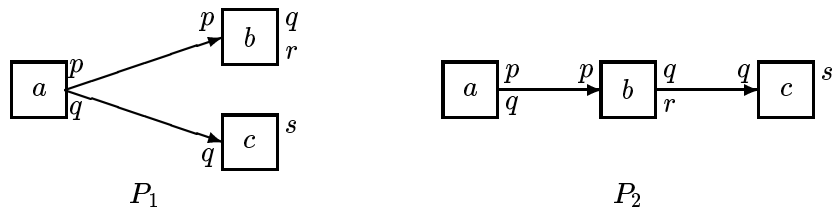

Figure 12: The p.o. plans in the failure example for VPC.

Suppose a total-order planner is given the PPI $\Pi = \langle \emptyset, \{r, s\} \rangle$ as input. It may then return either of the $\Pi$-valid t.o. plans $\langle a, b, c \rangle$ and $\langle a, c, b \rangle$, with action conditions as shown in Figure 12. When used as input to VPC, these two t.o. plans will give quite different results—the plan $\langle a, c, b \rangle$ will be converted to the p.o. plan $P_1$ in Figure 12, while the plan $\langle a, b, c \rangle$ will be converted to the p.o. plan $P_2$ in Figure 12. That is, in the first case VPC produces a plan which is not only a minimal-constrained deordering but even a minimum-constrained deordering, while in the second case it does not even produce a minimal-constrained deordering.[8]

The reason that VPC may fail to produce a minimal-constrained deordering is that it uses a non-admissible greedy strategy. Whenever it needs to find an operator $a$ achieving an effect required by the precondition of another operator $b$, it chooses the last such action ordered before $b$ in the input t.o. plan. However, there may be other actions earlier in the plan having the same effect and being a better choice.

## 8.2 The KK algorithm

Kambhampati and Kedar (1994) have presented an algorithm for generalising the ordering of a p.o. plan, using explanation-based generalisation. The algorithm is based on first constructing a validation structure for the plan and then use this as a guide in the generalisation phase. In the original paper, these computations are divided into two separate algorithms (EXP-MTC and EXP-ORD-GEN), but are here compacted into one single algorithm, KK (Figure 13). Furthermore, the version presented here is restricted to ground GT plans, while the original algorithm can also handle partially instantiated plans. This is no restriction for the results to be shown below.

The first part of the KK algorithm constructs a validation structure $\mathcal{V}$ for the plan, that is, an explanation for each precondition of every action in the plan. The validity criterion underlying this phase is a simplified version of Chapmans modal-truth criterion (Chapman,

---

8. Note that transitive arcs are omitted in the figures, so $P_2$ really has an ordering relation of size three. Although this example would not work if plans had been defined in the equally reasonable way that ordering relations should be intransitive, it is possible to construe similar examples also for this case.

```
1   procedure KK
2       Input: A valid self-contained p.o. plan ⟨A, ≺⟩
3       Output: A deordering of the input plan
4   comment Build a validation structure V for the plan
5   V ← ∅
6   Let ⟨a₁, . . . , aₙ⟩ be a topologically sorted version of ⟨A, ≺⟩
7   for 1 ≤ i ≤ n  do
8       for p ∈ pre(aᵢ)  do
9           Find min k < i s.t.
10              1. p ∈ post(aₖ) and
11              2. there is no j s.t.  k < j < i and ¬p ∈ post(aⱼ)
12          Add ⟨aₖ, p, aᵢ⟩ to V
13  comment Construct a generalised ordering ≺' for the plan
14  for each ⟨a, b⟩ ∈ ≺  do
15      Add ⟨a, b⟩ to ≺' if either of the following holds
16          1. a = aᵢ or a = a_G
17          2. ⟨a, p, b⟩ ∈ V for some p
18          3. ⟨c, p, a⟩ ∈ V and ¬p ∈ post(b)
19          4. ⟨b, p, c⟩ ∈ V and ¬p ∈ post(a)
20  return ⟨A, ≺'⟩
```

Figure 13: The KK algorithm

1987) without white knights. Since the algorithm is simplified to only handle ground plans here, an explanation is a causal link $\langle a_P, p, a_C \rangle$, meaning that the action $a_P$ produces the condition $p$ which is consumed by the action $a_C$. The algorithm constructs exactly one causal link for each precondition, and it chooses the earliest producer of $p$ preceeding $a_C$ with no intervening action producing $\neg p$ between this producer and $a_C$. The second phase of the algorithm builds a generalised ordering $\prec'$ for the plan based on this validation structure. To put things simply, only those orderings of the original plan are kept which either correspond to a causal link in the validation structure or that is required to prevent a threatening action to be unordered wrt. the actions in such a causal link.

It turns out that also the KK algorithm fails in generating plans that are guaranteed to be even minimal-constrained deorderings. Consider the t.o. plan $\langle A, B, C, D \rangle$ with action conditions as indicated in Figure 14. This t.o. plan is valid for the PPI $\langle \emptyset, \{r, s, t, u\} \rangle$. Since the KK algorithm always chooses the earliest possible producer of a precondition for the validation structure, it will build the validation structure $\{\langle A, p, D \rangle, \langle A, s, a_G \rangle, \langle B, q, D \rangle, \langle B, t, a_G \rangle, \langle C, r, a_G \rangle, \langle D, u, a_G \rangle\}$. Hence, the final ordering produced by KK will be as shown in Figure 14a. However, this plan is not a minimal-constrained deordering of the original plan, since it can be further deordered as shown in Figure 14b and remain valid. In this example, the input plan was totally ordered. In the case of partially ordered input plans, the behaviour of the algorithm depends on the particular topological order chosen. So the algorithm may or may not find a minimal-constrained deordering, but it is impossible to guarantee that it will succeed for all plans. Similarly, the authors mention that one may consider different ways of constructing the validation struc-ture. This would clearly also modify the behaviour and it remains an open question whether it is possible to generate, in polynomial time, a validation structure that guarantees that a minimal-constrained deordering is constructed in the second phase of the algorithm. Finding a validation structure that guarantees a minimum-constrained deordering is obviously an NP-hard problem since the second phase of the algorithm is polynomial.

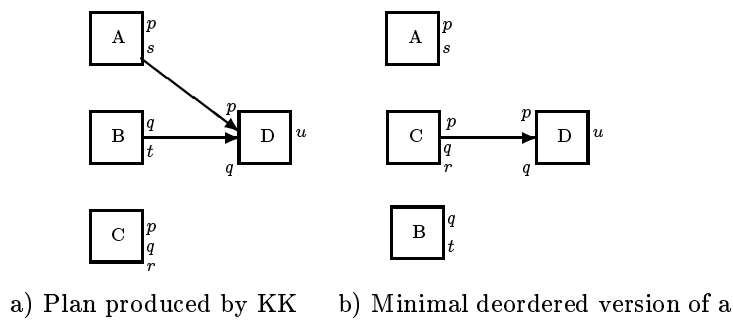

a) Plan produced by KK    b) Minimal deordered version of a

Figure 14: Failure example for the KK algorithm

## 8.3 Planners with Optimality Guarantees

The planning algorithm GRAPHPLAN (Blum & Furst, 1997) has a notion of time steps and tries to pack as many non-interacting actions as possible into one single time step. Furthermore, GRAPHPLAN finds the shortest plan, using the number of time steps as the measure. If assuming unit time and that all actions considered as non-interacting by GRAPHPLAN can be executed in parallel, then there is no plan having a shorter parallel execution than the plan produced by GRAPHPLAN. That is, GRAPHPLAN produces minimum reordered parallel plans under these assumptions. The second assumption is no limitation in practice, since each non-concurrency relation can be encoded by introducing a new atom and letting one of the interacting actions add it while the other one deletes it. The unit time assumption is more serious, however, especially since this assumption is likely not to hold in most applications. In the car-assembly scenario in Section 2, for instance, GRAPHPLAN would produce a plan that corresponds to the plan in Figure 5. Hence, the plan produced under the unit-time assumption happens to coincide with the optimal plan when taking actual execution times into account. This is just a fortunate coincidence, however, depending on the particular durations of actions in this example. Suppose instead that the durations of the actions are slightly different such that PAC has duration 2 and MvT1 has duration 8. Then the plan produced by GRAPHPLAN, which corresponds to the plan in Figure 5, does not have a faster schedule than 19 time units. This is not optimal since the plan in Figure 4 can be scheduled to execute in 17 time units for these particular duration times. Furthermore, it must be remembered that GRAPHPLAN is anyway restricted to those cases where a GT-equivalent planning language is sufficient, although recent improvements extend it to somewhat more expressive languages (Gazen & Knoblock, 1997; Köhler, Nebel, Hoffman, & Dimopoulos, 1997).

Knoblock (1994) has modified the UCPOP planner with a resource concept which makes it avoid unordered interacting actions. This means that the resulting planner produces definite parallel plans. Knoblock further modified the evaluation heuristic of the search to take parallel execution time into account. It thus seems as if this planner might be able to produce minimum reordered parallel plans, but the paper does not provide sufficient details to determine whether this is the case. It is also unclear whether the heuristic can handle actions with different duration times.

Yet another example is the polynomial-time planner for the $SAS^+$-IAO planning language (Jonsson & Bäckström, 1998) which produces plans which are minimum-constrained reordered. That is, for this restricted formalism it is clearly possible to optimise the ordering in polynomial time.

## 9. Discussion

The previous section listed a few planning algorithms from the literature that produce or attempt to produce plans which are least constrained or minimum parallel reordered. They do so only under certain restrictions, though. Furthermore, plans are not always generated 'from scratch', but can also be generated by modifying some already existing plan, referred to as case-based planning, or by repairing a plan that has failed during the execution phase. In such cases, the old plan may contain many ordering relations that will be obsolete in the modified/repaired plan. In fact, the KK algorithm (Kambhampati & Kedar, 1994) is motivated in the context of case-based planning. It is also important to remember that today, and probably for a long time into the future, very few plans are generated entirely by computer programs. The vast majority of plans in various applications are designed by humans, possibly with computer support. Already for quite small plans, it is very difficult for a human to see whether the ordering constraints are optimal or not, so computer support for such analyses is vital for designing optimal plans. For the same reason, also hierarchical-task-network planners, eg O-PLAN (Currie & Tate, 1991) and SIPE (Wilkins, 1988), produce plans where reordering actions could lead to better schedules. Such a planner often commits to one of the two possible orderings for a pair of actions based on expert-knowledge rules. However, it is hardly possible for a human expert to design rules that in all situations will guarantee that the optimal ordering choice is made.

On the coarseness level of complexity analysis it does not matter whether the tasks of planning, plan optimization and scheduling are integrated or separated since the total resulting complexity will be the same in both cases—the latter two computations are at most NP-complete and will, thus, be dominated by the planning, which is PSPACE-complete or worse. However, for good reasons this has not prevented the research community from studying planning and scheduling as separate problems, since understanding each problem in isolation also helps understanding the overall process. For the same reason, it is important to also study separately the problems discussed and analysed in this article. Furthermore, on a more fine-grained, practical level there might be considerable differences in efficiency between integrating the three computations and doing them separately. For instance, even if all three computations take exponential time, each of the problems considered in isolation may have fewer parameters, in which case it may be much more efficient to solve them in isolation. On the other hand, solving the whole problem at once may make it easier to do global optimisation. Which is the better will depend both on which methods are used and on various properties of the actual application, and it seems unlikely that one of the methods should always be the better.

As has been shown in this article, minimum reordering is a much better optimality criterion than minimum deordering, if only considering the overall parallel execution time. However, this is not necessarily true if also considering further metric constraints for subsequent scheduling. Deordering a plan can only add to the number of feasible schedules, while reordering may also remove some or, in the worst case, all feasible schedules. On the other hand, reordering may also lead to new and better schedules not reachable via deordering. Deordering can thus be viewed as a safe and, sometimes, cheap way to allow for better schedules, while reordering is an expensive method which has a potential for generating considerably better plans, but which may also make things worse. If using reordering in practice in cases where also metric scheduling constraints are involved, it seems necessary to use feedback from the scheduler to control the reordering process, or to try other reorderings. One could imagine a reordering algorithm which uses either heuristic search or randomized local-search methods à la GSAT (Selman, Levesque, & Mitchell, 1992) to find reorderings and then use the scheduler as evaluation function for the proposed reorderings.

While the plan modifications studied in this article may add considerably to the optimizations that are possible with traditional scheduling only, there is still a further potential of optimization left to study—modifying not only the action order, but also the set of actions. Such modification is already done in plan adaptation, but then only for generating a new plan from old cases, and optimizations in the sense of this article are not considered. Some preliminary studies of action-set modifications appear in the literature, though. Fink and Yang (1992) study the problem of removing redundant actions from total-order plans, defining a spectrum of redundancy criteria and analysing the complexity of achieving these. It is less clear that it is interesting to study action addition; adding actions to a plan could obviously not improve the execution time of it if it is to be executed sequentially. However, in the case of parallel execution of plans it has been shown that adding actions to a plan can sometimes allow for faster execution (Bäckström, 1994). Finally, if allowing both removal and addition of actions, an even greater potential for optimising plans seems available, but this problems seems not yet studied in the literature.

## 10. Conclusions

This article studies the problem of modifying the action ordering of a plan in order to optimise the plan according to various criteria. One of these criteria is to make a plan less constrained and the other is to minimize its parallel execution time. Three candidate definitions are proposed for the first of these criteria, constituting a spectrum of increasing optimality guarantees. Two of these are based on deordering plans, which means that ordering relations may only be removed, not added, while the last one builds on reordering, where arbitrary modifications to the ordering are allowed. The first of the three candidates, subset-minimal deordering, is tractable to achieve, while the other two, deordering or re-ordering a plan to minimize the size of the ordering, are both NP-hard and even difficult to approximate.

Similarly, optimising the parallel execution time of a plan is studied both for deordering and reordering of plans. In the general case, both of these computations are NP-hard and difficult to approximate. However, based on an algorithm from the literature it is shown that optimal deorderings can be computed in polynomial time for definite plans for a class of planning languages based on the notions of producers, consumers and threats, which includes most of the commonly used planning languages. Computing optimal reorderings can potentially lead to even faster parallel executions, but this problem remains NP-hard and difficult to approximate even under quite severe restrictions. Furthermore, deordering a plan is safe with respect to subsequent scheduling, while reordering a plan may remove feasible schedules, making deordering a good, but often suboptimal, approach in practice.

## Acknowledgements

Tom Bylander, Thomas Drakengren, Mark Drummond, Alexander Horz, Peter Jonsson, Bernhard Nebel, Erik Sandewall, Sylvie Thibeaux and the anonymous referees provided helpful comments on this article and previous versions of it. The research was supported by the Swedish Research Council for Engineering Sciences (TFR) under grants Dnr. 92-143 and 95-731.

## Appendix A

**Theorem 7.10** MINIMUM PARALLEL REORDERING *remains NP-hard even when restricted to total-order GT plans with only toggling unary actions and under the assumption of unit time, simple concurrency and that no actions are redundant.*

**Proof:** Proof by reduction from 3SAT (Garey & Johnson, 1979, p. 259). Let $\mathcal{P} = \{p_1, \ldots, p_n\}$ be a set of atoms and $C = \{C_1, \ldots, C_m\}$ a set of clauses over $\mathcal{P}$ s.t. for $1 \leq i \leq m$, $C_i = \{l_{i,1}, l_{i,2}, l_{i,3}\}$ is a set of three literals over $\mathcal{P}$.

First define the set of atoms

$$\mathcal{Q} = \{p_i^F, p_i^T, q_i \mid 1 \leq i \leq n\} \cup \{c_{i,j}, r_{i,j} \mid 1 \leq i \leq n, 1 \leq j \leq 3\}.$$

Then define a GT PPI $\Pi = \langle I, G \rangle$ with initial and goal states defined as

$$
\begin{aligned}
I &= Neg(\mathcal{Q}) \\
G &= \{p_i^F, p_i^T, \neg q_i \mid 1 \leq i \leq n\} \cup \{c_{i,j}, \neg r_{i,j} \mid 1 \leq i \leq n, 1 \leq j \leq 3\}
\end{aligned}
$$

Also, for each atom $p_i \in \mathcal{P}$, define four actions according to Table 2.
Further, for each clause $C_i \in C$, define nine actions according to Table 3 where

$$l_{i,j}^* = \begin{cases} p_k^F & \text{if } l_{i,j} = \neg p_k \\ p_k^T & \text{if } l_{i,j} = p_k. \end{cases}$$

Let $A$ be the set of all $4n + 9m$ actions thus defined. Clearly there is some total order $\prec$ s.t. the plan $P = \langle A, \prec \rangle$ is $\Pi$-valid. It is also obvious that none of the actions is redundant.

It is a trivial observation that any parallel execution $r$ of any $\Pi$-valid reordering of $P$ must satisfy that for each $i$, $1 \leq i \leq n$, either

$$r(A_i^F) < r(A_i^+) < r(A_i^T) < r(A_i^-)$$

or

$$r(A_i^+) < r(A_i^T) < r(A_i^-) < r(A_i^F),$$

and for each $i$, $1 \leq i \leq m$,

$$r(C_{i,k_1}^+) < r(B_{i,k_1}^+) < \left\{ \begin{array}{c} r(C_{i,k_1}^-) \\ r(C_{i,k_2}^+) \end{array} \right\} < r(B_{i,k_2}^+) < \left\{ \begin{array}{c} r(C_{i,k_2}^-) \\ r(C_{i,k_3}^+) \end{array} \right\} < r(B_{i,k_3}^+) < r(C_{i,k_3}^-),$$

where $k_1, k_2, k_3$ is a permutation of the numbers $1, 2, 3$. (This is to be interpreted s.t. the actions $C_{i,k_1}^-$ and $C_{i,k_2}^+$ can be released in either order, or simultaneously, and analogously for the actions $C_{i,k_2}^-$ and $C_{i,k_3}^+$).

The remainder of this proof shall show that $P$ can be reordered to have a parallel execution of length 8 iff the set $C$ of clauses is satisfiable.

*if:* Suppose $C$ is satisfiable. Let $I$ be a truth assignment for the atoms in $\mathcal{P}$ that satisfies $C$. Wlg. assume $I(p_i) = T$ for all $i$. Further, for each clause $C_j$, let $l_j$ be any literal in $C_j$ which is satisfied by $I$. Disregarding the action order for a moment, choose a release-time function $r$ for the actions as follows. For $1 \leq i \leq n$, let

$$r(A_i^+) = 0, \ r(A_i^T) = 1, \ r(A_i^-) = 2, \ r(A_i^F) = 3.$$

Further, for each $j$, $1 \leq j \leq m$, choose $k_1$ s.t. $l_{j,k_1} \in C_j$ is satisfied by $I$ (at least one such choice must exist by the assumption). Let $l_{j,k_2}$ and $l_{j,k_3}$ be the remaining two literals in $C_j$. Assign release times s.t. for $1 \leq h \leq 3$,

$$r(C_{j,k_h}^+) = 2h - 1, \ r(B_{j,k_h}^+) = 2h \ , r(C_{j,k_h}^-) = 2h + 1.$$

Now define the partial order $\prec'$ on $A$ s.t. for all actions $a, b \in A$, $a \prec' b$ iff $r(a) < r(b)$. Clearly, the plan $\langle A, \prec' \rangle$ is a $\Pi$-valid reordering of $P$ and $r$ is a parallel execution of length 8 for $\langle A, \prec' \rangle$. (Note that no other choice of $I$ could force a longer execution, while there is an execution of length 7 in the case where $C$ is satisfied by setting all atoms false.)

| operator | precond. | postcond. |
|----------|----------|-----------|
| $A_i^F$ | $\neg p_i^F, \neg q_i$ | $p_i^F$ |
| $A_i^T$ | $\neg p_i^T, q_i$ | $p_i^T$ |
| $A_i^+$ | $\neg q_i$ | $q_i$ |
| $A_i^-$ | $q_i$ | $\neg q_i$ |

Table 2: Generic actions for each atom $p_i$ in the proof of Theorem 7.10.

| operator | precond. | postcond. |
|----------|----------|-----------|
| $B_{i,1}^+$ | $l_{i,1}^*, r_{i,1}, \neg r_{i,2}, \neg r_{1,3}, \neg c_{i,1}$ | $c_{i,1}$ |
| $B_{i,2}^+$ | $l_{i,2}^*, \neg r_{i,1}, r_{i,2}, \neg r_{1,3}, \neg c_{i,2}$ | $c_{i,2}$ |
| $B_{i,3}^+$ | $l_{i,3}^*, \neg r_{i,1}, \neg r_{i,2}, r_{1,3}, \neg c_{i,3}$ | $c_{i,3}$ |
| $C_{i,1}^+$ | $\neg r_{i,1}$ | $r_{i,1}$ |
| $C_{i,1}^-$ | $r_{i,1}$ | $\neg r_{i,1}$ |
| $C_{i,2}^+$ | $\neg r_{i,2}$ | $r_{i,2}$ |
| $C_{i,2}^-$ | $r_{i,2}$ | $\neg r_{i,2}$ |
| $C_{i,3}^+$ | $\neg r_{i,3}$ | $r_{i,3}$ |
| $C_{i,3}^-$ | $r_{i,3}$ | $\neg r_{i,3}$ |

Table 3: Generic atoms for each clause $C_i$ in the proof of Theorem 7.10.

*only if:* Suppose $C$ is not satisfiable. Further suppose that $Q$ is a minimum reordering of $P$ and that $r$ is a parallel execution of length 8 or shorter for $Q$. Wlg. assume that every action is released as early as possible by $r$. Then, according to the observation above it must hold for each $i$, $1 \le i \le n$, that either

$$r(A_i^F) = 0, \ r(A_i^+) = 1, \ r(A_i^T) = 2, \ r(A_i^-) = 3$$

or

$$r(A_i^+) = 0, \ r(A_i^T) = 1, \ r(A_i^-) = 2, \ r(A_i^F) = 3.$$

Hence, exactly one of the atoms $p_i^F$ and $p_i^T$ is true at time 2. Let $p_i^*$ denote this atom. Since $r$ is of length 8, it follows from the earlier observation that for all $j$, $1 \le j \le m$, $r(B_{j,k}^+) \le 2$ for some $k$, $1 \le k \le 3$. Hence, $l_{j,k} = p_i^*$ for some $i$, since $Q$ is $\Pi$-valid and $r$ is a parallel execution for $Q$. Define an interpretation $I$ s.t. for all $i$, $1 \le i \le n$,

$$I(p_i) = \begin{cases} F, & \text{if } p_i^* = p_i^F \\ T, & \text{otherwise}. \end{cases}$$

However, this interpretation is obviously a model for $C$, which contradicts the assumption. It follows that $r$ must be of length 9 or longer.

This concludes the proof and shows that $C$ is satisfiable iff $P$ has a reordering with a parallel execution of length 8 or not. $\qquad\square$

**Theorem 7.11** MINIMUM PARALLEL DEORDERING *cannot approximate* MINIMUM PARALLEL REORDERING *within* $|A|^k$ *for any constant* $k \ge 0$.

**Proof:** The proof assumes GT plans and simple concurrency. First, define the generic actions $a_i^k(m)$, $b_i^k$ and $c_i^k(m)$ according to Table 10. Further, define recursively the generic plans

$$P_i^k(m) = \begin{cases} \langle a_{(i-1)m+1}^1(1), b_{(i-1)m+1}^0, c_{(i-1)m+1}^1(1), \ldots, a_{im}^1(1), b_{im}^0, c_{im}^1(1) \rangle, & \text{for } k = 1 \\ \langle a_{(i-1)m+1}^k(m); P_{(i-1)m+1}^{k-1}(m); c_1^k(m), \ldots, a_{im}^k(m); P_{im}^{k-1}(m); c_{im}^k(m) \rangle, & \text{for } k > 1. \end{cases}$$

Furthermore, for arbitrary $k, n > 0$ define the PPI $\Pi_n^k = \langle \{p_1^k, \ldots, p_n^k\}, \{q_1^k, \ldots, q_n^k\} \rangle$.

Now, prove the claim that for arbitrary $k, n > 0$, the plan $P_1^k(n)$

1. is $\Pi_n^k$-valid,

2. has no deordering of length less than $3n^k + \sum_{i=1}^{k-1} 2n^i$ and

3. has a reordering of length $2k + 1$.

Proof by induction over $k$.

*Base case (k=1):* Choose an arbitrary $n > 0$. The plan $P_1^1(n)$ is obviously $\Pi_n^k$-valid and has no deordering other than itself, which is of length $3n$. Consider the reordering $Q_1^1(n)$ of $P_1^1(n)$ with the same actions and with ordering relation $\prec$ defined s.t. for all $i$, $1 \leq i \leq n$, $a_i^1(1) \prec b_i^0 \prec c_i^1(1)$ and for all $i$, $1 < i \leq n$, $a_i^1(1) \prec b_{i-1}^0$. This reordering is $\Pi^k(n)$-valid and has a parallel execution $r_1^1(n)$ of length 3, defined s.t. for all $i$, $1 \leq i \leq n$, $r_1^1(n)(a_i^1(1)) = 1$, $r_1^1(n)(b_i^0) = 2$ and $r_1^1(n)(c_i^1(1)) = 3$. (This plan is shown in Figure 15.) The claim is thus satisfied for the base case.

*Induction:* Suppose the claim is satisfied for all $l < k$, for some $k \geq 1$ and prove that the claim holds also for $l = k$. Choose an arbitrary $n > 0$. It follows from the induction hypothesis that none of the subplans $P_1^{k-1}(n) \ldots, P_n^{k-1}(n)$ can be deordered, so they have to remain totally ordered. Furthermore, for all $i$, $1 \leq i \leq n$, it is necessary that the action $a_i^k(n)$ is ordered before the subplan $P_i^{k-1}(n)$ and that the action $c_i^k(n)$ is ordered after it. It is also clear that for no $i$, $1 \leq i \leq n$ can the order $c_i^k(n) \prec a_{i+1}^k(n)$ be removed without making the plan invalid. Hence, $P_1^k(n)$ has no other deordering than itself, which is of length

$$\sum_{i=1}^{n} (2 + length(P_i^{k-1}(n)) = n(2 + length(P_1^{k-1}(n)))$$

$$= 2n + n(3n^{k-1} + \sum_{i=1}^{k-2} 2n^i) = 3n^k + \sum_{i=1}^{k-1} 2n^i,$$

which proves the deordering case of the claim.

For the reordering case, define a reordering $Q_1^k(n)$ of $P_1^k(n)$ with the same actions and with ordering relation defined as follows. For each subplan $P_i^{k-1}(n)$ of $P_1^k(n)$, reorder its actions so it has length $2(k-1) + 1$, which is possible according to the induction hypothesis. Further, for each $i$, $1 \leq i \leq n$, and each $j$, $(i-1)n + 1 \leq j \leq in$ order $a_i^k(n) \prec a_j^{k-1}(n)$ and $c_j^{k-1}(n) \prec c_i^k(n)$ (or $a_{i+1}^k(n) \prec a_j^{k-1}(1)$ and $c_j^{k-1}(1) \prec c_i^k(n)$ for the case $k = 2$). Hence, each segment of the type $a_i^k(n); P_i^{k-1}(n); c_i^k(n)$ is reordered to have length $2k + 1$. Finally, for each $i$, $1 \leq i \leq n$, order $a_i^k(n) \prec a_{(i-1)n}^{k-1}(n)$ (or $a_i^k(n) \prec a_{(i-1)n}^{k-1}(1)$ for the case $k = 2$). The plan $Q_1^k(n)$ is $\Pi^k(n)$-valid since the subplans $P_1^{k-1}(n), \ldots, P_n^{k-1}(n)$ do not have any atoms in common and, thus, the $\#$ relation does not hold between any two actions belonging to different such subplans. This reordered plan can be executed under the parallel execution $r_i^k(n)$ defined s.t. $r_i^k(n)(a_i^k(n)) = 1$, $r_i^k(n)(c_i^k(n)) = 2k + 1$ and for all $i$, $1 \leq i \leq n$ and all actions $a' \in Q_i^{k-1}(n)$, $r_i^k(n)(a') = r_i^{k-1}(n)(a') + 1$. Since this is a parallel execution of length $2k + 1$ for the reordered plan, the claim holds also for $k$.

| action | pre-condition | post-condition |
|--------|---------------|----------------|
| $a_i^k(m)$ | $\{p_i^k\}$ | $\{p_{(i-1)m+1}^{k-1}, \ldots, p_{im}^{k-1}, \neg q_{(i-1)m}^{k-1}\}$ |
| $b_i^k$ | $\{p_i^k\}$ | $\{q_i^k\}$ |
| $c_i^k(m)$ | $\{q_{(i-1)m+1}^{k-1}, \ldots, q_{im}^{k-1}\}$ | $post(c_i^k(m)) = \{q_i^k\}$. |

Table 4: Generic actions for the proof of Theorem 7.11.

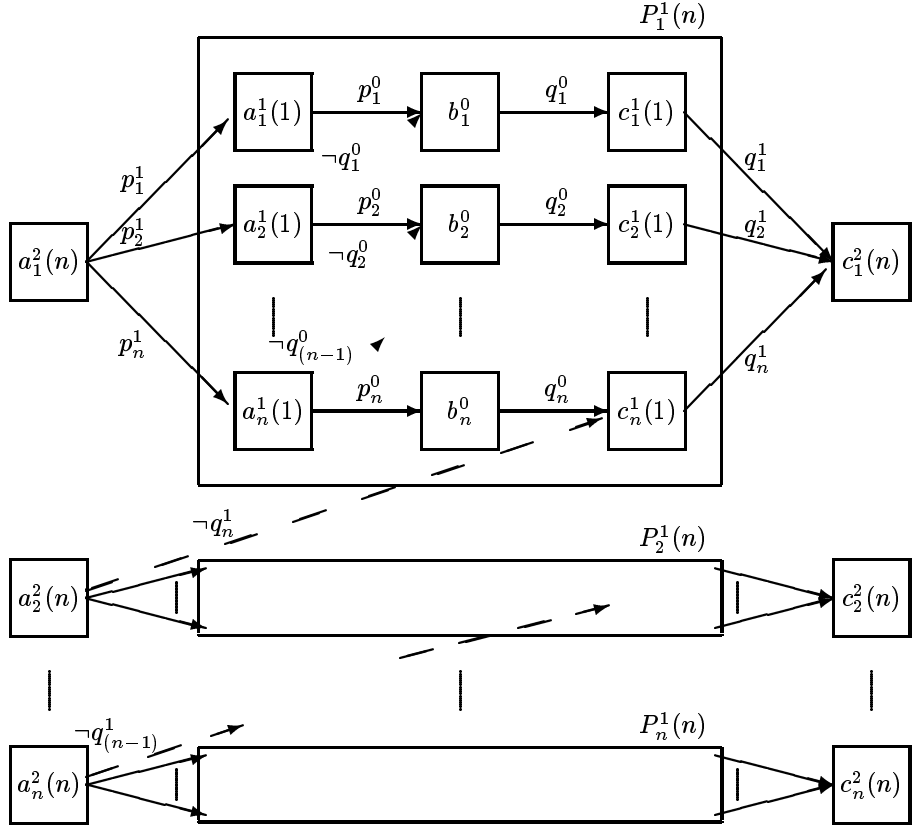

Figure 15: The reordering $Q_1^2(n)$ of the plan $P_1^2(n)$ as an example of the induction case in the proof of Theorem 7.11 (solid arrows denote orderings required by producer-consumer relationships and are labelled with the atom produced/consumed, while dashed arrows denote ordering constraints to avoid threats and are labelled with the possibly conflicting atom).

This concludes the induction, so the claim holds for all $k > 0$. Since

$$\frac{3n^k + \sum_{i=1}^{k-1} 2n^i}{2k + 1} \geq \frac{1}{(2k+1)3^{k-1}} |A|^k$$

for all $k > 0$, the theorem holds. □